# Interpretable Style Takagi-Sugeno-Kang Fuzzy Clustering

Suhang Gu[1,2], Ye Wang[1], Yongxin Chou[2], Jinliang Cong[2], Mingli Lu[2], and Zhuqing Jiao[1]

[1] School of Wangzheng Microelectronics, Changzhou University, Changzhou 213164, People's Republic of China
[2] the School of Electrical Engineering and Automation, Changshu Institute of Technology, Suzhou
215500, People's Republic of China
(Corresponding author: Zhuqing Jiao)

**Abstract.** Clustering is an efficient and essential technique for exploring latent knowledge of data. However, limited attention has been given to the interpretability of the clusters detected by most clustering algorithms. In addition, due to the homogeneity of data, different groups of data have their own homogeneous styles. In this paper, the above two aspects are considered, and an interpretable style Takagi-Sugeno-Kang (TSK) fuzzy clustering (IS-TSK-FC) algorithm is proposed. The clustering behavior of IS-TSK-FC is fully guided by the TSK fuzzy inference on fuzzy rules. In particular, samples are grouped into clusters represented by the corresponding consequent vectors of all fuzzy rules learned in an unsupervised manner. This can explain how the clusters are generated in detail, thus making the underlying decision-making process of the IS-TSK-FC interpretable. Moreover, a series of style matrices are introduced to facilitate the consequents of fuzzy rules in IS-TSK-FC by capturing the styles of clusters as well as the nuances between different styles. Consequently, all the fuzzy rules in IS-TSK-FC have powerful data representation capability. After determining the antecedents of all the fuzzy rules, the optimization problem of IS-TSK-FC can be iteratively solved in an alternation manner. The effectiveness of IS-TSK-FC as an interpretable clustering tool is validated through extensive experiments on benchmark datasets with unknown implicit/explicit styles. Specially, the superior clustering performance of IS-TSK-FC is demonstrated on case studies where different groups of data present explicit styles. The source code of IS-TSK-FC can be downloaded from https://github.com/gusuhang10/IS-TSK-FC.

## 1 INTRODUCTION

Clustering, an important unsupervised learning technique [1], has been applied in many fields, such as pattern recognition [2], image segmentation [3], and data mining [4]. By solving clustering problems, the inherent structure and meaningful knowledge of different types of data can be explored. When encountering clustering problems in complex real-world applications such as security, privacy, and intelligent medical diagnosis, researchers are always hoped to provide detailed explanations of their recommendations [5-6]. In addition, the focus of most clustering algorithms is efficient cluster assignment, including aspects such as high accuracy and robustness to noisy data, while few works have focused on cluster interpretability. Thus, users cannot easily comprehend how clusters are generated by clustering algorithms and how each cluster differs from the other clusters.

Moreover, samples from the same group usually occur as a pattern, and share the homogeneous style because of the homogeneity of data [17-20]. Typical examples include art style recognition [21-22], with different paintings having unique styles such as Cubism or Post impressionism, as shown in Fig. 1(a); fashion style recognition [23-24], with clothing images divided into different groups according to clothing styles such as bohemian or goth, as shown in Fig. 1(b); and architectural style recognition [24-25], with buildings classified by considering architectural styles such as Georgian or Queen Anne architecture, as shown in Fig. 1(c). In machine learning, conventional algorithms usually assume that samples are identically and independently distributed (i.i.d.) [19,25]. However, styles of data are not i.i.d., which is an important aspect of style-based algorithms that distinguishes them from conventional algorithms [19,25].

### 1.1 Existing Works on Interpretable Clustering

To facilitate the comprehensibility of clustering analysis, there have been some related studies which can be mainly grouped into two groups. The first type addresses the interpretable clustering by introducing the decision tree method [5-9]. Since the original decision tree is a supervised approach, it needs to be converted to an unsupervised approach [5]. Then, each cluster can be represented by one or several leaf nodes [5-9]. Moreover, based on the unsupervised decision tree, the path from the root node to a leaf node can be regarded as a rule [8-9]. With these two characteristics of the first type of clustering algorithm, how an instance is classified into a cluster can be explained, and the constructed clusters are thus inherently interpretable.

The second type aims to generate interpretable clustering results by introducing fuzzy rules [10-15]. For example, in [10], a rapid fuzzy rule clustering algorithm was proposed, which described input features using IF-THEN fuzzy rules and performed clustering by iteratively conducting granular computing. In [11], a fuzzy rule-based clustering (FRBC) algorithm was proposed.

FRBC takes the inputs as the main data and simultaneously generates uniformly distributed samples as auxiliary data. Then, the two kinds of data are regarded as two classes with assigned labels, and the cluster analysis is transformed into a fuzzy rule-based classification task. The strategy of converting the unsupervised clustering problem into a supervised classification problem can also be found in [12]. In [13] and [14], two fuzzy clustering algorithms were proposed for accurate clustering in a wireless sensor network. The former used predefined fuzzy rules to optimize the adjustment of the competition radius for the cluster head, which is the leader in each cluster, while the latter focused on determining a genetic fuzzy output function for each fuzzy rule by modifying the clonal selection algorithm. In [15], a novel Takagi-Sugeno-Kang (TSK) fuzzy clustering (TSK-FC) algorithm was proposed. One of the characteristics of TSK-FC is that its clustering behavior is directly guided by fuzzy inference on TSK fuzzy rules.

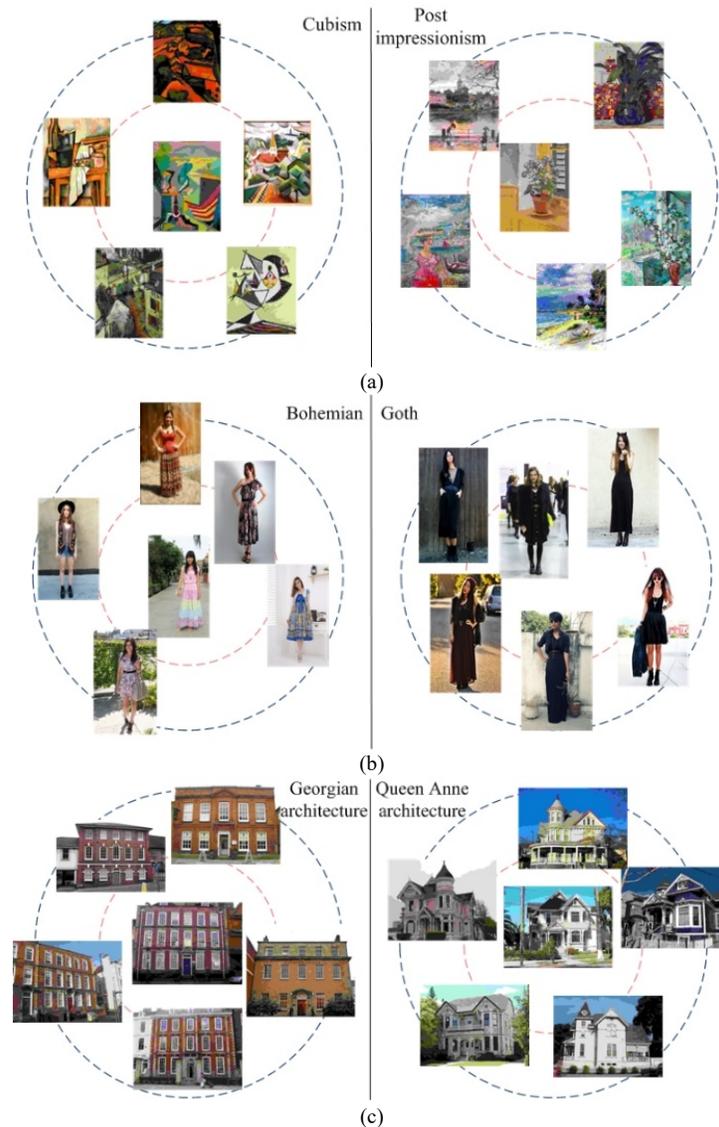

**Fig. 1** Typical examples of styles of data including (a) art style recognition, (b) fashion style recognition, and (c) architectural style recognition.

Compared to the first type of clustering algorithm, the second type of clustering algorithm is more likely to obtain transparent clustering results because fuzzy rule-based algorithms use fuzzy sets with linguistic terms to conduct fuzzy inference, and fuzzy rules are inherently explainable [16]. However, with the exception of TSK-FC [15], the fuzzy rules introduced by the second type of clustering algorithm are used only as intermediate tools for supporting cluster analysis, which means that the corresponding clustering process is not fully guided by fuzzy inference on fuzzy rules. This leads to limited cluster interpretability and is one of the two main problems to be solved in this paper. For TSK-FC, since its whole clustering process is completely dominated by fuzzy inference on fuzzy rules, comprehensible clustering results can be obtained [15]. Thus, we focus on TSK-FC in this paper.

## 1.2 Existing Works on Style-based Algorithms

Some works have attempted to utilize the styles of data for efficient learning, and they can be grouped into three types: 1) style descriptor-based algorithms, 2) statistical dependence-based algorithms, and 3) style matrix-based algorithms. According to [21-24,26-28,30], the first type of algorithm represents the styles of data by generating style descriptors. In particular, discriminative style representations of data can be obtained using deep learning methods, such as autoencoders [24], convolution neural networks (CNNs) [21,28], ensemble models [26], and generative adversarial networks (GANs) [27,30]. Generally, the excellent performance of the first type of algorithm relies on accurate extraction of style descriptors. According to [17,25,29], since the styles of data break the i.i.d. assumption, the second type of algorithm describes the styles of data based on statistical dependences. Specifically, each style of data is defined as a style probability that covers a broad range of probability distributions. One of the main distinctions of the second type of algorithm is that the styles can be directly estimated from different groups of data. However, this type of algorithm requires that the distributions of the styles show significant differences [17].

The third type of algorithm aims to learn the styles of data using a series of matrices which are termed style matrices [18-19,31-35]. In [18] and [19], field Bayesian models and field support vector machines were proposed by introducing style matrices into Bayesian decision theory and support vector machine (SVM), respectively. Similarly, in [32] and [33], the prototypical transfer learning and TSK fuzzy classifier were extended based on style matrices and the corresponding regularization terms. Consequently, the derived style transfer learning and style TSK fuzzy classifier can adjust style variations between the source and target domains and generate fuzzy rules with powerful representation capabilities. Different from the supervised style-based algorithms in [18], [19], [32] and [33], unsupervised style-based clustering algorithms [31,34-35] have also been developed, which explore the styles of data using style matrices and are very important in this paper. Please note that the style matrices used in all style matrix-based algorithms are square except those used in the fuzzy style flat-based clustering [35], where the style matrices for considering physical features (e.g., distance, color, or similarity) and stylistic features simultaneously are not square.

Compared to the first and second types of style learning algorithms, the style matrices used in the third type of algorithm are more flexible for exploring style information of data [18-19,32]. Moreover, the corresponding regularization terms can be used to adjust the styles of data and capture the nuances between different styles [18-19,31-35]. Therefore, we use style matrices to solve another important problem associated with learning the styles of data, which facilitates the design of an interpretable style clustering algorithm.

## 1.3 Motivations and Contributions

In this paper, we consider both interpretable cluster analysis and styles of data and aim to develop a novel interpretable style clustering algorithm, which is motivated by reviewing the existing works on interpretable clustering [5-15] and style-based algorithms [17-35]. Developing an interpretable style clustering is a nontrivial task because the interpretability of the whole clustering procedure must be guaranteed and samples with the same style must be grouped into the same cluster. To this end, we first present the following considerations.

1) An interpretable style clustering algorithm should perform well on data with explicit styles, e.g., the data presented in Fig. 1. However, when clustering data with implicit styles, namely, data that may or may not have styles, the designed interpretable style clustering algorithm should function as a basic clustering tool that can achieve comparable performance at least to conventional clustering algorithms.

2) J. M. Mendel et al. [16] claimed that "*fuzzy models or, more specifically, fuzzy rule-based models, can easily be understood and interpreted by a human user or data analyst.*" In addition, the TSK fuzzy inference system (FIS) is more interpretable and precise than other fuzzy rule-based systems [15-16,33,36,37]. Consequently, we select the TSK fuzzy model as the starting point for developing the interpretable style clustering model and intend to realize a transparent clustering procedure with the designed clustering model based on IF-THEN fuzzy rules.

3) Based on previous works on style-based algorithms [18-19,31-35], introducing style matrices to explore the styles of data to improve clustering performance should not affect the interpretability of the proposed interpretable style clustering algorithm. Moreover, the introduced style matrices should become part of the IF-THEN fuzzy rules rather than independent components to enable careful inspection of the designed clustering model.

Based on the above considerations, a novel interpretable style Takagi-Sugeno-Kang fuzzy clustering (IS-TSK-FC) algorithm is proposed. IS-TSK-FC is derived from the conventional TSK FIS, which is usually used to solve classification or regression problems [15-16,33,36,37,39]; however, IS-TSK-FC considers the styles of data simultaneously. Similar to the conventional TSK FIS, IS-TSK-FC can obtain interpretable clustering results via fuzzy inference on TSK fuzzy rules. In addition, the samples in the same cluster generated by IS-TSK-FC have homogeneous styles, which are different from those of the samples in the other clusters.

The main contributions of this paper can be summarized as follows.

1) In the clustering process, the clustering behavior of the proposed IS-TSK-FC algorithm is completely guided by fuzzy inference on fuzzy rules rather than by considering fuzzy rules as a supplementary measure for addressing clustering problems [10-14]. Moreover, once the antecedents of all fuzzy rules are determined, each generated cluster corresponds to its own consequent vector rather than all generated clusters corresponding to only one consequent vector of all fuzzy rules. Thus, the process of grouping different samples into clusters using IS-TSK-FC is transparent, and the decision-making process of IS-TSK-FC is interpretable.

2) By considering the styles of data, samples with the same style are grouped into the same cluster. Moreover, a regularization term for the styles of data is introduced into the objective function of IS-TSK-FC. With this regularization term and the corresponding regularization parameter, the styles of different clusters can be explored, and specific style information in each style matrix corresponding to each cluster can be mathematically calculated to reflect detailed nuances between different styles of data. This improves the discriminability of IS-TSK-FC for samples with different styles.

3) The proposed IS-TSK-FC algorithm is applicable in interpretable cluster analysis, and the consequent parts of all fuzzy rules are learned in an unsupervised manner, which is significantly different from the learning process of a conventional TSK FIS, in which the consequent parts of all fuzzy rules are learned in a supervised manner [15-16,33,36, 37]. In addition, due to the introduction of style matrices, the consequent parts of all fuzzy rules in IS-TSK-FC contain both conventional consequent parameters (as seen in (1)) and style matrices (as seen in (15)). Consequently, a novel type of fuzzy rule is designed for IS-TSK-FC, and each fuzzy rule has powerful representation ability due to the style matrices. Moreover, after determining the antecedents of all fuzzy rules, the consequent vectors and style matrices of the consequent parts of all fuzzy rules can be alternatively determined by referring to [38].

### 1.4 Outline of This Paper

The remainder of this paper is organized as follows. In Section II, the conventional TSK fuzzy inference system [36] is briefly described. In Section III, the proposed interpretable style TSK fuzzy clustering IS-TSK-FC algorithm is described in detail, and in Section IV, extensive experimental results of IS-TSK-FC are reported. Section V concludes this paper.

## 2 TSK Fuzzy Inference System

Since the TSK FIS [36] is selected as the starting point for developing IS-TSK-FC, we first briefly review the TSK FIS. As one of the most widely used fuzzy rule-based inference systems, the TSK FIS enables uncertain and linguistic terms, such as high, low, fast, and slow, to be represented using fuzzy logic, thus resembling human reasoning [37]. The TSK FISs can achieve a good balance between precision and interpretability [37]. The TSK FISs include a series of IF-THEN fuzzy rules, where the antecedent part is formed by fuzzy sets, and the output is determined by a linear function of the input features in the consequent part. For an input $\mathbf{x} = [x_1, x_2, \cdots, x_d]^T \in \Re^{d \times 1}$, where $d$ denotes the dimensionality, the $r$th fuzzy rule $ru^r$ of a TSK FIS can be defined as follows:

$$ru^r: \text{IF } x_1 \text{ is } A_1^r \ \wedge \ x_2 \text{ is } A_2^r \ \wedge \cdots \wedge \ x_d \text{ is } A_d^r$$
$$\text{THEN } f^r(\mathbf{x}) = p_0^r + p_1^r x_1 + \cdots + p_d^r x_d, 1 \le r \le R \tag{1}$$

where $R$ denotes the number of fuzzy rules and $A_i^r (1 \le i \le d)$, corresponding to the $r$th fuzzy rule, denotes the fuzzy subset for the $i$th feature $x_i$. Additionally, $\wedge$ denotes the fuzzy conjunction operator, and $p_0^r, p_1^r, \cdots, p_d^r$ are the consequent parameters in the consequent part of the $r$th fuzzy rule. Here, when the consequent part includes only $p_0^r$, i.e., $f^r(\mathbf{x}) = p_0^r$, the TSK FIS is termed a zero-order TSK FIS. If the consequent part is a linear function of $p_0^r, p_1^r, \cdots, p_d^r$, i.e., $f^r(\mathbf{x}) = p_0^r + p_1^r x_1 + \cdots + p_d^r x_d$, the TSK FIS is termed a first-order TSK FIS. After determining the antecedents of each fuzzy rule and applying a defuzzification strategy, the output of the TSK FIS for input $\mathbf{x}$ can be computed as follows:

$$y^o = \sum_{r=1}^{R} \frac{\mu^r(\mathbf{x})}{\sum_{r'=1}^{R} \mu^{r'}(\mathbf{x})} f^r(\mathbf{x}) = \sum_{r=1}^{R} \tilde{\mu}^r(\mathbf{x}) f^r(\mathbf{x}) \tag{2}$$

where $\mu^r(\mathbf{x})$ represents the firing strength of the $r$th fuzzy rule [39]. In addition, $\tilde{\mu}^r(\mathbf{x})$ denotes the normalized version of $\mu^r(\mathbf{x})$. For $\mu^r(\mathbf{x})$, this term can be defined as

$$\mu^r(\mathbf{x}) = \prod_{i=1}^{d} \mu_{A_i^r}(x_i). \tag{3}$$

According to [39], $\mu_{A_i^r}(\mathbf{x})$ is usually expressed as the Gaussian membership function, which is defined as

$$\mu_{A_i^r}(x_i) = \exp\left(-\frac{1}{2}\left(\frac{x_i - c_i^r}{\delta_i^r}\right)^2\right) \tag{4}$$

where $c_i^r$ and $\delta_i^r$ denote the center and width of the Gaussian membership function, respectively, and can be determined by clustering techniques [37]. For example, if fuzzy c-means (FCM) clustering [40] is used, $c_i^r$ and $\delta_i^r$ can be estimated as

$$c_i^r = \sum_{j=1}^{N} \mu_{jr} x_{ji} \bigg/ \sum_{j=1}^{N} \mu_{jr} \tag{5}$$

and

$$\delta_i^r = h \sum_{j=1}^{N} \mu_{jr} \left( x_{ji} - c_i^r \right) \bigg/ \sum_{j=1}^{N} \mu_{jr}, \tag{6}$$

respectively. In (4) and (5), $N$ denotes the number of inputs, and the FCM can be used to obtain $\mu_{jr}$, which denotes the membership degree of the $j$th input $\mathbf{x}_j$ with respect to the $r$th cluster. In addition, $h$ denotes the scale parameter, which can be set using learning strategies such as cross-validation [37]. For clarity, other membership functions can also be used to construct TSK FISs for specific modeling tasks.

After determining the antecedent parts of all the fuzzy rules for the TSK FIS, we can define the following transformations.

$$\mathbf{x}_e = \left[ 1, \mathbf{x}^T \right]^T, \tag{7}$$

$$\tilde{\mathbf{x}}^r = \tilde{\mu}^r (\mathbf{x}) \mathbf{x}_e, \tag{8}$$

$$\mathbf{x}_g = \left[ \left( \tilde{\mathbf{x}}^1 \right)^T, \left( \tilde{\mathbf{x}}^2 \right)^T, \cdots, \left( \tilde{\mathbf{x}}^R \right)^T \right]^T, \tag{9}$$

$$\mathbf{p}^r = \left[ p_0^r, p_1^r, \cdots, p_d^r \right]^T, \tag{10}$$

$$\mathbf{P}_g = \left[ \left( \mathbf{p}^1 \right)^T, \left( \mathbf{p}^2 \right)^T, \cdots, \left( \mathbf{p}^R \right)^T \right]^T. \tag{11}$$

Based on (7)–(11), the output $y^o$ of the TSK FIS in (2) can be further expressed as

$$y^o = \mathbf{P}_g^T \mathbf{x}_g \tag{12}$$

where $\mathbf{P}_g \in \mathfrak{R}^{R(1+d) \times 1}$ is the combined consequent vector of all fuzzy rules. Consider an input dataset $\mathbf{X} = \left[ \mathbf{x}_1, \mathbf{x}_2, \cdots, \mathbf{x}_N \right]^T \in \mathfrak{R}^{N \times d}$ with the label set $\mathbf{Y} = \left[ y_1, y_2, \cdots, y_N \right]^T \in \mathfrak{R}^{N \times 1}$. Referring to [41], the following criterion function can be minimized:

$$\min_{\mathbf{P}_g} E = \sum_{i=1}^{N} \left( y_i - \mathbf{P}_g^T \mathbf{x}_{gi} \right)^2 = \left( \mathbf{Y} - \mathbf{X}_g \mathbf{P}_g \right)^T \left( \mathbf{Y} - \mathbf{X}_g \mathbf{P}_g \right) \tag{13}$$

where $\mathbf{x}_{gi}$, corresponding to the $i$th input $\mathbf{x}_i$, is determined using (9) and $\mathbf{X}_g = \left[ \mathbf{x}_{g1}, \mathbf{x}_{g2}, \cdots, \mathbf{x}_{gN} \right]^T \in \mathfrak{R}^{N \times R(1+d)}$ denotes the antecedent matrix. The least square solution of $\mathbf{P}_g$ can be determined as

$$\mathbf{P}_g = \left( \mathbf{X}_g^T \mathbf{X}_g \right)^{-1} \mathbf{X}_g^T \mathbf{Y}. \tag{14}$$

Please note that other learning strategies, such as ridge regression or regularized least squares regression [42-44], can also be considered to determine the consequent vector $\mathbf{P}_g$.

# 3 The Proposed Model

In this section, the proposed interpretable style TSK fuzzy clustering IS-TSK-FC is explained in detail. On the one hand, the whole clustering procedure of IS-TSK-FC is dominated by fuzzy inference on fuzzy rules. On the other hand, samples with the homogeneous style are grouped into the same cluster by IS-TSK-FC. Based on these two characteristics, we attempt to design a novel type of TSK fuzzy rule.

## 3.1 On IS-TSK-FC

Consider an input dataset $\mathbf{X} = \left[ \mathbf{x}_1, \mathbf{x}_2, \cdots, \mathbf{x}_N \right]^T \in \mathfrak{R}^{N \times d}$ with $N$ samples and assign these samples to $K$ clusters. As stated above, we apply the advantages of the TSK FIS [16,33,36-37,39] and style-based algorithms [17-35] to cluster analysis and propose a novel IS-TSK-FC for interpretable style fuzzy clustering. Here, style matrices are introduced to represent the style information of each cluster, which also serve as the consequents of each fuzzy rule. Thus, a novel type of fuzzy rule is designed for IS-TSK-FC, which has the following form (taking the $r$th fuzzy rule $Ru^r$ as an example):

$$Ru^r : \text{IF } x_1 \text{ is } A_1^r \ \wedge \ x_2 \text{ is } A_2^r \ \wedge \cdots \wedge \ x_d \text{ is } A_d^r$$
$$\text{THEN } \mathbf{f}^r (\mathbf{x}_i) = \left[ f_1^r (\mathbf{x}_i), f_2^r (\mathbf{x}_i), \cdots, f_K^r (\mathbf{x}_i) \right]. \tag{15}$$

In (15), $1 \le i \le N$, $1 \le r \le R$, and $\mathbf{f}^r(\mathbf{x}_i)$ is the consequent part of the $r$th fuzzy rule in the form of a $K$-dimensional vector, where the $k$th element $f_k^r(\mathbf{x}_i) = \left(\mathbf{p}_k^r\right)^T \mathbf{s}_k^r \mathbf{x}_e^i \left(1 \le k \le K\right)$ corresponds to the $k$th cluster. Besides, $\mathbf{s}_k^r \in \mathfrak{R}^{(1+d)\times(1+d)}$ denotes the $k$th style matrix associated with the $r$th fuzzy rule. $\mathbf{x}_e^i$, associated with the $i$th sample $\mathbf{x}_i$, can be determined using (7)–(9). $\mathbf{p}_k^r = \left[p_{0,k}^r, p_{1,k}^r, \cdots, p_{d,k}^r\right]^T$ consists of the conventional consequents, as in (1). Based on (15), each fuzzy rule in IS-TSK-FC is used to map the fuzzy set $\left[A_1^r, A_2^r, \cdots, A_d^r\right]$ to the $K$-dimensional vector $\mathbf{f}^r(\mathbf{x}_i)$, and by applying a simple defuzzification strategy [37], we can obtain the following $K$-dimensional output of IS-TSK-FC:

$$\mathbf{y}_i^o = \mathbf{F}(\mathbf{x}_i)\tilde{\mathbf{\mu}}(\mathbf{x}_i) \tag{16}$$

where $\mathbf{F}(\mathbf{x}_i) = \left[\mathbf{f}^1(\mathbf{x}_i), \mathbf{f}^2(\mathbf{x}_i), \cdots, \mathbf{f}^R(\mathbf{x}_i)\right]^T$ and $\tilde{\mathbf{\mu}}(\mathbf{x}_i) = \left[\tilde{\mu}^1(\mathbf{x}_i), \tilde{\mu}^2(\mathbf{x}_i), \cdots, \tilde{\mu}^R(\mathbf{x}_i)\right]^T$. For clarity, the $r$th element $\tilde{\mu}^r(\mathbf{x}_i)$ in $\tilde{\mathbf{\mu}}(\mathbf{x}_i)$ is the same as in (2). Moreover, based on $\mathbf{p}_k^r$ in each fuzzy rule in $f_k^r(\mathbf{x}_i)$ in (15), analogous to (11), the $k$th consequent vector of all fuzzy rules can be expressed as

$$\mathbf{P}_{gk} = \left[\left(\mathbf{p}_k^1\right)^T, \left(\mathbf{p}_k^2\right)^T, \cdots, \left(\mathbf{p}_k^R\right)^T\right]^T \tag{17}$$

which corresponds to the $k$th cluster. Therefore, the $K$-dimensional output of IS-TSK-FC in (16) can be further expressed as

$$\mathbf{y}_i^o = \left[\mathbf{P}_{g1}^T \mathbf{S}_1 \mathbf{x}_{gi}, \mathbf{P}_{g2}^T \mathbf{S}_2 \mathbf{x}_{gi}, \cdots, \mathbf{P}_{gK}^T \mathbf{S}_K \mathbf{x}_{gi}\right]^T \tag{18}$$

where $\mathbf{x}_{gi} \in \mathfrak{R}^{R(1+d)\times 1}$, associated with $\mathbf{x}_i$, is determined using (7)–(9), and $\mathbf{P}_{gk} \in \mathfrak{R}^{R(1+d)\times 1}$. Moreover, $\mathbf{S}_k \in \mathfrak{R}^{R(1+d)\times R(1+d)}$ is the style matrix of the $k$th cluster, which consists of $\mathbf{s}_k^r (1 \le r \le R)$ of each fuzzy rule in $f_k^r(\mathbf{x}_i)$ in (15). Similar to the TSK FIS [36-37], when only $p_0^r$ is included in $\mathbf{p}_k^r$ for the consequent part $\mathbf{f}^r(\mathbf{x}_i)$, the IS-TSK-FC is termed the zero-order IS-TSK-FC. If $\mathbf{p}_k^r = \left[p_{0,k}^r, p_{1,k}^r, \cdots, p_{d,k}^r\right]^T$ in the consequent part $\mathbf{f}^r(\mathbf{x}_i)$, the IS-TSK-FC is termed the first-order IS-TSK-FC.

According to (18), once $\left\{\mathbf{P}_{gk}\right\}$ and $\left\{\mathbf{S}_k\right\}$ are provided, the $i$th sample $\mathbf{x}_i$ can be grouped into the cluster associated with the smallest value in $\mathbf{y}_i^o$. However, it is nontrivial to determine $\left\{\mathbf{P}_{gk}\right\}$ and $\left\{\mathbf{S}_k\right\}$ because IS-TSK-FC is an unsupervised learning technique. Thus, all the learning strategies of problem (13) which require the real labels of samples input into the TSK FIS [41-44] are no longer suitable for IS-TSK-FC.

To address the optimization problem of $\left\{\mathbf{P}_{gk}\right\}$ and $\left\{\mathbf{S}_k\right\}$, we consider the following. According to [45], fuzzy inference on the IF-parts of all fuzzy rules can be regarded as evaluating the kernel function in kernel machines such as SVMs [19]. This indicates that fuzzy inference on the IF-parts of all fuzzy rules equivalently transforms the original feature space of the samples into a high-dimensional feature space; thus, an optimal kernel function does not need to be identified, leading to good generalizability for the fuzzy rule-based system. Consequently, we solve the optimization problem of $\left\{\mathbf{P}_{gk}\right\}$ and $\left\{\mathbf{S}_k\right\}$ in a high-dimensional feature space via fuzzy inference on the IF-parts of all fuzzy rules for IS-TSK-FC.

For IS-TSK-FC with $R$ fuzzy rules, after determining the antecedents of all fuzzy rules based on (3)–(6) and forming the corresponding antecedent matrix $\mathbf{X}_g = \left[\mathbf{x}_{g1}, \mathbf{x}_{g2}, \cdots, \mathbf{x}_{gN}\right]^T \in \mathfrak{R}^{N\times R(1+d)}$, where each $\mathbf{x}_{gi} \in \mathfrak{R}^{R(1+d)\times 1} (1 \le i \le N)$ associated with $\mathbf{x}_i$ can be determined using (7)–(9), we consider calculating $K$ consequent vectors of all fuzzy rules in the $R(1+d)$-dimensional feature space to determine the appropriate cluster for $\mathbf{x}_i$. This process can be written as

$$\mathbf{P}_{gk}^T \mathbf{x}_{gi} + q_k = 0, 1 \le k \le K \tag{19}$$

where $\mathbf{P}_{gk}$ is determined in an unsupervised manner in this paper, and $q_k \in \mathfrak{R}$ is the bias. For clarity, when determining the antecedents of all fuzzy rules, the FCM [40] is adopted to determine the center and width of the Gaussian membership function in (4). Furthermore, the introduced style matrices are the consequents of all fuzzy rules, as shown in (15). Thus, problem (19) can be further expressed as

$$\mathbf{P}_{gk}^T \mathbf{S}_k \mathbf{x}_{gi} + q_k = 0, 1 \le k \le K. \tag{20}$$

Based on (20), our aim is to represent each cluster by the corresponding consequent vector of all fuzzy rules rather than all clusters corresponding to only one consequent vector in the TSK FIS, as shown in (12) and (13). Moreover, we aim to group samples with the same style into the same cluster. Thus, we need to solve the following programming problems:

$$\min_{\mathbf{P}_{gk}, q_k, \mathbf{S}_k} \left\| \mathbf{X}_{gk} \mathbf{S}_k \mathbf{P}_{gk} + q_k \mathbf{e}_k \right\|_2^2 + \lambda \left\| \mathbf{S}_k - \mathbf{I} \right\|_F^2$$
$$\text{s.t. } \left\| \mathbf{P}_{gk} \right\|_2^2 = 1, 1 \le k \le K \tag{21}$$

where $\mathbf{X}_{gk}$ denotes the antecedent matrix corresponding to subset $\mathbf{X}_k \in \mathfrak{R}^{N_k \times d} \subseteq \mathbf{X}$, which belongs to the $k$th cluster; $N_k$ denotes the number of samples in $\mathbf{X}_k$; $\mathbf{e}_k \in \mathfrak{R}^{N_k \times 1}$ denotes a vector of ones; $\lambda$ denotes a regularization parameter; $\mathbf{I} \in \mathfrak{R}^{R(1+d) \times R(1+d)}$ denotes the identity matrix; and the constraint normalizes the consequent vector of each cluster. In addition, $\|\bullet\|_2$ and $\|\bullet\|_F$ denote the $L_2$ and Frobenius norms, respectively.

The interpretation of problem (21) is clear. First, in addition to learning the consequent vectors $\{\mathbf{P}_{gk}\}$ of all fuzzy rules for each cluster, a series of style matrices $\{\mathbf{S}_k\}$ for all clusters are jointly learned. In addition, each style matrix $\mathbf{S}_k$ is exploited to mathematically describe the style information in each cluster. Second, the Frobenius term $\lambda \|\mathbf{S}_k - \mathbf{I}\|_F^2$ is used as the regularization term, which prevents the style matrices from being identity matrices and simultaneously penalizes overly flexible style information for each cluster. Moreover, the regularization parameter $\lambda$ has a significant impact on the exploration of the styles of different clusters. Specifically, lower values of $\lambda$ lead to overly flexible style information, which results in the clustering performance of IS-TSK-FC being mainly dominated by the styles of data. However, if $\lambda$ is set to $+\infty$, each style matrix $\mathbf{S}_k$ tends to be an identity matrix $\mathbf{I}$. With no styles of data applied, IS-TSK-FC then becomes the TSK-FC [15]. Consequently, based on the regularization term $\lambda \|\mathbf{S}_k - \mathbf{I}\|_F^2$ in problem (21), the style information of each cluster can be appropriately adjusted, thereby allowing IS-TSK-FC to capture the nuances between different styles and effectively discriminate samples with different styles.

Once the solutions of $\{\mathbf{P}_{gk}\}$ and $\{\mathbf{S}_k\}$ are obtained by solving problem (21), all the samples are reassigned to the clusters according to their decision values as

$$y_i^o = \arg \min_{k=1,2,\cdots,K} \left| \mathbf{P}_{gk}^T \mathbf{S}_k \mathbf{x}_{gi} + q_k \right|. \tag{22}$$

Moreover, the assignments of all the samples and the solutions of $\{\mathbf{P}_{gk}\}$ and $\{\mathbf{S}_k\}$ are updated in an alternating manner until a repeat overall assignment appears or the maximum number $H$ of alternating updates is met.

### 3.2 Optimization

As indicated in (15), the consequent part of each fuzzy rule in IS-TSK-FC contains both the consequent vectors $\{\mathbf{P}_{gk}\}$ and style matrices $\{\mathbf{S}_k\}$. Here, the optimization problem of IS-TSK-FC in (21) can be divided into two independent suboptimization problems: consequent vector and style matrix suboptimization problems. Moreover, by appropriately initializing the style matrices $\{\mathbf{S}_k\}$, i.e., $\mathbf{S}_k = \mathbf{I}, 1 \leq k \leq K$, an alternating optimization strategy [19,38] can be employed to solve the two suboptimization problems to identify sufficient local minima for $\{\mathbf{P}_{gk}\}$ and $\{\mathbf{S}_k\}$ [19].

1) *Consequent vector suboptimization problem:* By fixing the style matrices $\{\mathbf{S}_k\}, 1 \leq k \leq K$, the optimization problem of IS-TSK-FC in (21) is transformed as:

$$\min_{\mathbf{P}_{gk}, q_k} \left\| \mathbf{X}_{gk} \mathbf{S}_k \mathbf{P}_{gk} + q_k \mathbf{e}_k \right\|_2^2 \tag{23}$$
$$\text{s.t. } \left\| \mathbf{P}_{gk} \right\|_2^2 = 1, 1 \leq k \leq K.$$

By referring to [46-47], we find that the solutions of $\{\mathbf{P}_{gk}\}$ for problem (23) can be obtained by solving $K$ eigenvalue problems in the $R(1+d)$-dimensional feature space, which is determined by fuzzy inference on the IF-parts of the $R$ fuzzy rules. Specifically, by applying the Lagrangian multiplier method [47], we construct the Lagrangian function of (23), which is expressed as

$$L_1 = \left\| \mathbf{X}_{gk} \mathbf{S}_k \mathbf{P}_{gk} + q_k \mathbf{e}_k \right\|_2^2 - \xi_k \left( \left\| \mathbf{P}_{gk} \right\|_2^2 - 1 \right) \tag{24}$$

where $\xi_k$ denotes the Lagrangian multiplier. By setting the partial derivatives of $L_1$ with respect to both $\mathbf{P}_{gk}$ and $q_k$ to zero, we can obtain the following two equalities:

$$\frac{1}{2} \frac{\partial L_1}{\partial \mathbf{P}_{gk}} = \mathbf{S}_k^T \mathbf{X}_{gk}^T \left( \mathbf{X}_{gk} \mathbf{S}_k \mathbf{P}_{gk} + q_k \mathbf{e}_k \right) - \xi_k \mathbf{P}_{gk} = 0, \tag{25}$$

$$\frac{1}{2} \frac{\partial L_1}{\partial q_k} = \mathbf{e}_k^T \left( \mathbf{X}_{gk} \mathbf{S}_k \mathbf{P}_{gk} + q_k \mathbf{e}_k \right) = 0. \tag{26}$$

Based on the above two equalities, we can further obtain

$$\mathbf{S}_k^T \mathbf{X}_{gk}^T \mathbf{X}_{gk} \mathbf{S}_k \mathbf{P}_{gk} + \mathbf{S}_k^T \mathbf{X}_{gk}^T q_k \mathbf{e}_k = \xi_k \mathbf{P}_{gk} \tag{27}$$

$$q_k = -\frac{\mathbf{e}_k^T \mathbf{X}_{gk} \mathbf{S}_k \mathbf{P}_{gk}}{\mathbf{e}_k^T \mathbf{e}_k}. \tag{28}$$

By substituting $q_k$ in (27) with that in (28), we can obtain

$$\mathbf{H}_k \mathbf{P}_{gk} = \xi_k \mathbf{P}_{gk} \tag{29}$$

where $\mathbf{H}_k$ is expressed as

$$\mathbf{H}_k = \mathbf{S}_k^T \mathbf{X}_{gk}^T \mathbf{X}_{gk} \mathbf{S}_k - \frac{\mathbf{S}_k^T \mathbf{X}_{gk}^T \mathbf{e}_k \mathbf{e}_k^T \mathbf{X}_{gk} \mathbf{S}_k}{\mathbf{e}_k^T \mathbf{e}_k}. \tag{30}$$

Thus, according to [46-47], $\mathbf{P}_{gk}$ can be easily determined as the eigenvector corresponding to the smallest eigenvalue of matrix $\mathbf{H}_k$ in (30). With the solution of $\mathbf{P}_{gk}$, $q_k$ can be determined using (28).

2) *Style matrix suboptimization problem:* When $\mathbf{P}_{gk}$ and $q_k$ $(1 \le k \le K)$ are fixed, the optimization problem of IS-TSK-FC in (21) can be written as

$$\min_{\mathbf{S}_k} \left\| \mathbf{X}_{gk} \mathbf{S}_k \mathbf{P}_{gk} + q_k \mathbf{e}_k \right\|_2^2 + \lambda \left\| \mathbf{S}_k - \mathbf{I} \right\|_F^2. \tag{31}$$

The solutions of $\mathbf{S}_k (1 \le k \le K)$ cannot be directly obtained by solving problem (31) using the Lagrangian multiplier method [47]. However, by referring to [38], we can solve problem (31) and determine $\mathbf{S}_k (1 \le k \le K)$ via an intermediate variable $\boldsymbol{\alpha}_k$, which is expressed as

$$\boldsymbol{\alpha}_k = \mathbf{X}_{gk} \mathbf{S}_k \mathbf{P}_{gk} + q_k \mathbf{e}_k \tag{32}$$

where each $\mathbf{S}_k$ is initially set to an identity matrix. Then, problem (31) can be further expressed as

$$\min_{\mathbf{S}_k} \boldsymbol{\alpha}_k^T \left( \mathbf{X}_{gk} \mathbf{S}_k \mathbf{P}_{gk} + q_k \mathbf{e}_k \right) + \lambda \left\| \mathbf{S}_k - \mathbf{I} \right\|_F^2. \tag{33}$$

With the Lagrangian multiplier method [47], the Lagrangian function of (33) is expressed as

$$L_2 = \boldsymbol{\alpha}_k^T \left( \mathbf{X}_{gk} \mathbf{S}_k \mathbf{P}_{gk} + q_k \mathbf{e}_k \right) + \lambda \left\| \mathbf{S}_k - \mathbf{I} \right\|_F^2, \tag{34}$$

and by setting the partial derivative of $L_2$ with respect to $\mathbf{S}_k$ to zero, we can obtain the following equation:

$$\frac{\partial L_2}{\partial \mathbf{S}_k} = \mathbf{X}_{gk}^T \boldsymbol{\alpha}_k \mathbf{P}_{gk}^T + 2\lambda \left( \mathbf{S}_k - \mathbf{I} \right) = 0. \tag{35}$$

Thus, each $\mathbf{S}_k$ is expressed as

$$\mathbf{S}_k = \mathbf{I} - \frac{1}{2\lambda} \mathbf{X}_{gk}^T \boldsymbol{\alpha}_k \mathbf{P}_{gk}^T. \tag{36}$$

Based on both (32) and (36), an iterative update procedure is applied between $\boldsymbol{\alpha}_k$ and $\mathbf{S}_k (1 \le k \le K)$. Therefore, each $\mathbf{S}_k$ is determined when it satisfies $\left\| \mathbf{S}_k^{t+1} - \mathbf{S}_k^t \right\|_F < \theta$, where $\theta$ denotes a termination threshold, $t (1 \le t \le T)$ denotes the $t$th iteration, and $T$ is the maximum number of iterations.

### 3.3 Implementation and Computational Complexity

After deriving the objective function of IS-TSK-FC and the corresponding optimization procedure, the algorithmic implementation of IS-TSK-FC is summarized in Algorithm 1.

The proposed interpretable style TSK fuzzy clustering IS-TSK-FC aims to partition samples directly based on fuzzy inference on fuzzy rules while simultaneously considering the styles of data. To this end, first, the labels of all samples need to be initialized using an existing clustering algorithm, such as $k$-means [4] or FCM [40]. We also need to calculate the antecedents of all fuzzy rules according to (3)–(9). For clarity, the FCM [40] is used to determine the center $c_i^r$ and width $\delta_i^r$ of the Gaussian membership function in (4) for each fuzzy rule. Then, after initializing each style matrix $\mathbf{S}_k$ as an identity matrix $\mathbf{I}$, the consequent vectors $\{\mathbf{P}_{gk}\}$, parameters $\{q_k\}$ and style matrices $\{\mathbf{S}_k\}$ are updated in an alternating manner. For clarity, the maximum number $H$ of alternating updates, maximum number $T$ of iterations and termination threshold $\theta$ in $\left\| \mathbf{S}_k^{t+1} - \mathbf{S}_k^t \right\|_F < \theta$ are set to 50, 30 and $10^{-3}$, respectively, which are appropriate values in most cases. According to Algorithm 1, the clustering procedure of IS-TSK-FC is completely based on fuzzy inference on fuzzy rules. As a result, the grouping of different samples into clusters can be explained, which makes the underlying decision-making process of IS-TSK-FC interpretable. In addition, as stated above, the optimization problem of IS-TSK-FC is divided into consequent vector and style matrix suboptimization problems. Moreover, these two suboptimization problems are solved by minimizing the specific objective functions in (23) and

(31) in each step when determining $\{\mathbf{P}_{gk}\}$ and $\{\mathbf{S}_k\}$. Generally, the optimization procedure of IS-TSK-FC is conducted in a monotonically descending manner until the algorithm reaches a stable state. This ensures that IS-TSK-FC converges with at least a local minimum for each $\mathbf{P}_{gk}$ or $\mathbf{S}_k (1 \le k \le K)$ [19].

---

**Algorithm 1:** Clustering procedure of IS-TSK-FC.

---

**Input:** Dataset $\mathbf{X} = [\mathbf{x}_1, \mathbf{x}_2, \cdots, \mathbf{x}_N]^T \in \Re^{N \times d}$, number $R$ of fuzzy rules, number $K$ of clusters, regularization parameter $\lambda$, maximum number $H$ of alternating updates, maximum number $T$ of iterations, and termination threshold $\theta$.

**Output:** The final consequent vectors $\{\mathbf{P}_{gk}\}$, parameters $\{q_k\}$, style matrices $\{\mathbf{S}_k\}$, and clustering label set.

**Procedure:**

1 Initialize the labels of all samples.

2 Set $\mathbf{S}_k = \mathbf{I} (1 \le k \le K)$.

3 Calculate the antecedents of all fuzzy rules using (3)–(9).

4 Set $h = 1$.

5 Repeat:

    **4**    **For** $k = 1; \ k \le K$ **do**//Determine $\{\mathbf{P}_{gk}\}$ and $\{\mathbf{S}_k\}$

    **5**    Determine $\mathbf{P}_{gk}$ as the eigenvector associated with the smallest eigenvalue of matrix $\mathbf{H}_k$ in (30).

    **6**    Calculate $q_k$ using (28).

    **7**    Set $t = 1$ and $\mathbf{S}_k^t = \mathbf{S}_k$.

    **8**    Repeat:

    **9**    Calculate $\boldsymbol{\alpha}_k^{t+1}$ using (32).

    **10**    Calculate $\mathbf{S}_k^{t+1}$ using (36).

    **11**    Until reaching $\left\| \mathbf{S}_k^{t+1} - \mathbf{S}_k^t \right\|_F < \theta$ and set $\mathbf{S}_k = \mathbf{S}_k^{t+1}$.

    **12**    **End**

13 $h = h + 1$.

14 Until a repeat overall assignment appears or the maximum number $H$ of alternating updates is met.

15 Output the final consequent vectors $\{\mathbf{P}_{gk}\}$, parameters $\{q_k\}$, style matrices $\{\mathbf{S}_k\}$, and clustering label set.

---

We next analyze the computational complexity of IS-TSK-FC. Algorithm 1 shows that the clustering procedure of IS-TSK-FC is dominated by solving the consequent vector and style matrix suboptimization problems. Therefore, we mainly describe the computational complexity based on Steps 5–14. Consider a dataset $\mathbf{X} = [\mathbf{x}_1, \mathbf{x}_2, \cdots, \mathbf{x}_N]^T \in \Re^{N \times d}$ where $N$ and $d$ denote the number of samples and dimensionality of each sample, respectively. Once the antecedents of the $R$ fuzzy rules are determined according to (3)–(9), we can obtain each antecedent matrix $\mathbf{X}_{gk}$ associated with the subset $\mathbf{X}_k \in \Re^{N_k \times d} \subseteq \mathbf{X}$, which belongs to the $k$th cluster, where $N_k$ denotes the number of samples in $\mathbf{X}_k$. Step 5 has a computational complexity of $O\left(R^3(1+d)^3\right)$; in this step, $\mathbf{P}_{gk}$ is determined as the eigenvector associated with the smallest eigenvalue of matrix $\mathbf{H}_k$ in (30). For Step 6, the computational complexity of calculating $q_k$ is approximately $O\left(N_k R(1+d) + R^2(1+d)^2\right)$. Additionally, the computational complexity of calculating $\boldsymbol{\alpha}_k^{t+1}$ using (32) in Step 9 is $O\left(N_k + N_k R^2(1+d)^2 + N_k R(1+d)\right)$, and the computational complexity of calculating $\mathbf{S}_k^{t+1}$ using (36) in Step 10 is $O\left(N_k R(1+d) + R^2(1+d)^2\right)$. Since the maximum number of iterations is $T$, the computational complexity of determining $\mathbf{S}_k$ is no more than $O\left(TN_k R^2(1+d)^2\right)$. Moreover, the for loop is executed $K$ times, the number $N_k$ of samples in each subset $\mathbf{X}_k$ is smaller than the number $N$ of total samples in $\mathbf{X}$, and the maximum number of alternating updates is $H$. Thus, the maximum computational complexity of Algorithm 1 is approximately $O\left(HK\left(R^3(1+d)^3 + TNR^2(1+d)^2\right)\right)$, which is linearly dependent on $N$, $R^3$ and $d^3$. This implies that IS-TSK-FC is suitable for solving clustering problems on the datasets where the data scales and dimensions are not large.

*Remark 1:* In [15], TSK fuzzy clustering TSK-FC was first designed to perform interpretable clustering directly based on fuzzy inference on TSK fuzzy rules. Compared to TSK-FC, the proposed IS-TSK-FC not only obtains fully interpretable clustering results but also has three distinct advantages. First, the styles of data are considered, and a series of style matrices and the corresponding regularization term are introduced into the objective function of IS-TSK-FC. Therefore, the styles of different clusters and the nuances between different styles can be mathematically exploited, which will be experimentally proven to be effective in helping IS-TSK-FC recognize samples with different styles. Second, the consequent part of each fuzzy rule contains both the consequent vectors and style matrices. This makes the fuzzy rules used in IS-TSK-FC more powerful for representing data than those used in TSK-FC. Third, the optimization problem of TSK-FC is solved via the concave–convex procedure (CCCP) [48], which results in the computational complexity linearly depending on $N^3$, $R^3$ and $d^3$ [15]. However, the optimization problem of IS-TSK-FC can be easily solved using an alternating update procedure, as shown in Algorithm 1, whose computational complexity linearly depends on $N$, $R^3$ and $d^3$. This implies that IS-TSK-FC has relatively lower computational complexity compared with TSK-FC.

### 3.4 Illustrative Example

To better understand the workflow of IS-TSK-FC, especially its interpretable decision-making process, which is an important property of a fuzzy system [5,36], Fig. 2 shows the mechanism of IS-TSK-FC based on an example with 504 samples, which can be grouped into four clusters, as shown in Fig. 2(a). For clarity, the number of fuzzy rules is set to 4, the regularization parameter $\lambda$ in (21) is set to $10^4$, and the FCM [40] is used to determine the center and width of the Gaussian membership function in (4).

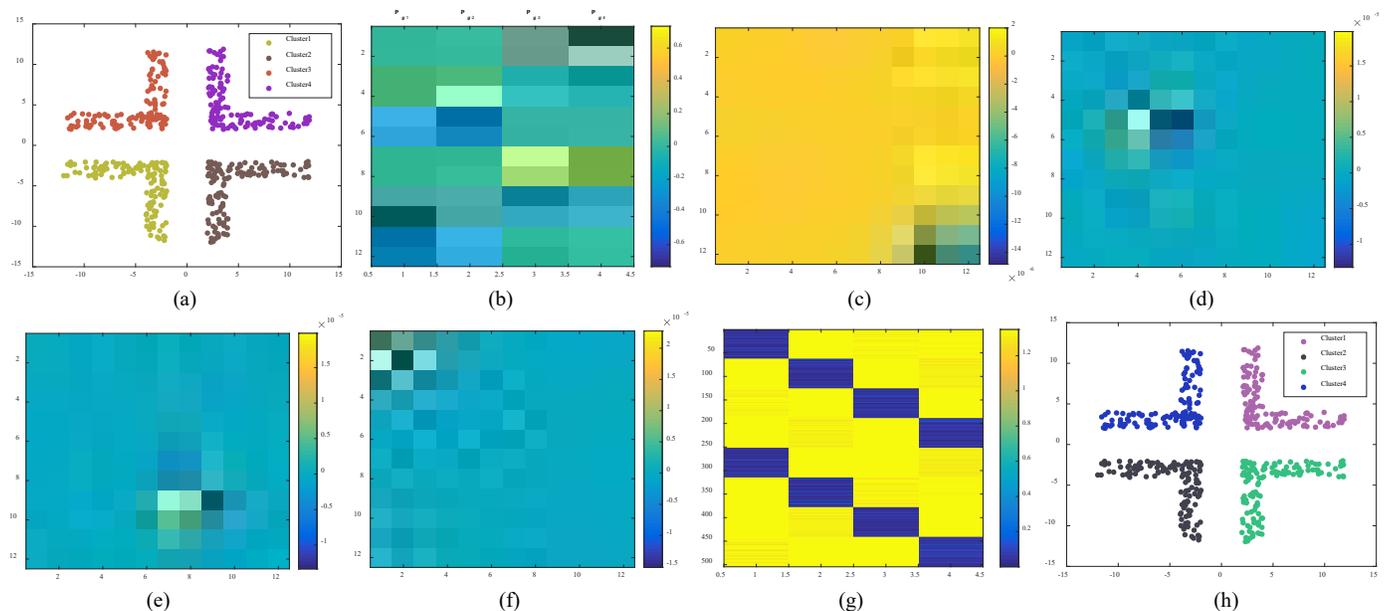

**Fig. 2** Illustration of IS-TSK-FC's mechanism. (a) The example with four clusters. (b) The visualization of the calculated consequent vectors $\mathbf{P}_{g1}$, $\mathbf{P}_{g2}$, $\mathbf{P}_{g3}$ and $\mathbf{P}_{g4}$ corresponding to four columns, respectively. (c)–(f) The visualization of the calculated style matrices $\mathbf{S}_1$, $\mathbf{S}_2$, $\mathbf{S}_3$ and $\mathbf{S}_4$ corresponding to the four clusters in (a), respectively. (g) The visualization of the decision values calculated using (22) for all samples in (a). (h) The final generated four clusters according to the decision values in (g) by IS-TSK-FC.

After determining the antecedents of the four fuzzy rules, four consequent vectors $\mathbf{P}_{g1}$, $\mathbf{P}_{g2}$, $\mathbf{P}_{g3}$, and $\mathbf{P}_{g4}$ associated with the four columns visualized in Fig. 2(b), are calculated and can be used to represent the four clusters in Fig. 2(h). Therefore, each cluster generated by IS-TSK-FC corresponds to its own consequent vector rather than all four generated clusters having the same consequent vector of all fuzzy rules. This substantially influences how the samples are grouped into the final four clusters, and the results can be directly interpreted through fuzzy inference on the four fuzzy rules applied by IS-TSK-FC. Besides, Fig. 2(c)–Fig. 2(f) visualize the four style matrices $\mathbf{S}_1$, $\mathbf{S}_2$, $\mathbf{S}_3$, and $\mathbf{S}_4$, respectively. For clarity, each pair of $\mathbf{P}_{gk}$ and $\mathbf{S}_k$ corresponds to only one cluster. In particular, Fig. 2(a) shows that the four clusters present similar shapes, which may indicate that the four clusters have similar styles. However, by inspecting Fig. 2(c)–Fig. 2(f), we find that the nuances (denoted by the blocks with different colors in different columns) among the styles of the four clusters can be captured through the regularization term in (21). This may facilitate the correct partition of all samples, which can be validated by the clustering results visualized in Fig. 2(h). Fig. 2(g) visualizes the decision values of all samples calculated using (22), and the results reflect demonstrate that all samples are correctly grouped into the appropriate clusters, with the clusters denoted by the blue rectangles. For clarity, each column in Fig. 2(g) corresponds to a cluster generated by IS-TSK-FC. Fig. 2(h) visualizes the final generated clusters according to the decision values of all samples shown in Fig. 2(g).

# 4 Experimental Results and Analysis

In this section, IS-TSK-FC is compared with several comparison algorithms, and extensive experimental results are reported to validate the effectiveness of IS-TSK-FC. In particular, the distinctive potential of IS-TSK-FC is demonstrated on the datasets with explicit styles. In Subsection IV.A, the details of the comparison algorithms and experimental setups are described. In Subsection IV.B, the feasibility of IS-TSK-FC as a basic clustering tool is examined based on fourteen benchmark datasets. In Subsections IV.C, the superiority of IS-TSK-FC is investigated in seven case studies.

### 4.1 Comparison Algorithms and Settings

Since the proposed IS-TSK-FC considers both the styles of data and the interpretability of the clustering results through fuzzy inference on fuzzy rules, the following clustering algorithms are chosen for comparison. The fuzzy subspace clustering (FSC) algorithm can detect regular subspace clusters by using fuzzy techniques for subspace clustering [57]. The FCM is a classic soft clustering technique based on fuzzy logic [40], and its improved version, i.e., FWCM_FCM [56], is based on feature weight and cluster weight learning. The fuzzy rule-based clustering (FRBC) algorithm explores clusters in data by transforming the unsupervised cluster analysis into a supervised learning problem solved using a fuzzy rule-based classification system [11]. The TSK-FC algorithm conducts a completely interpretable clustering procedure via direct fuzzy inference on fuzzy rules [15]. Here, two versions of TSK-FC, i.e., zero-order and first-order TSK-FCs, are adopted. For simplicity, they are termed 0-TSK-FC and 1-TSK-FC, respectively. The fuzzy style $k$-plane clustering algorithm, S-KPC, is based on twofold data representations, i.e., style matrices for representing style information of clusters and enhanced nodes for improved data discrimination [31]. The two versions of IS-TSK-FC, i.e., zero-order and first-order IS-TSK-FCs, are described in Subsection III.A and termed 0-IS-TSK-FC and 1-IS-TSK-FC, respectively, for simplicity. In addition, the detailed parameter settings of all the comparison algorithms are summarized in Table 1.

**Table 1** Detailed parameter settings of all comparison algorithms

| Algorithms | Parameter settings |
| --- | --- |
| FSC | The weight component $\alpha$ is set to 2, and the parameter $\varepsilon$ is set to $10^{-5}$ by following [57]. |
| FCM | All involved parameters are set by following [58]. |
| FWCW_FCM | By following [56], the parameter $\alpha$ is set to 2, the parameter $p$ is adjusted in the range of 0 to 0.5 with an interval of 0.01, the threshold value $\varepsilon$ is set to $10^{-5}$, the memory effect $\beta$ is tuned over $\{0,0.1,\cdots,0.9,1.0\}$, the exponent of the attribute weight $q$ is set to 2, and the fuzzy degree is set to 2. |
| FRBC | The threshold $\tau$ is set to 0.1 by following [11], and a heuristic method is used to generate a suitable value for the prespecified number of $Q$ rules per class according to [59]. |
| 0-TSK-FC | Gaussian membership functions are chosen to calculate the antecedents of all fuzzy rules, and the associated centers and kernel widths are set according to [60]. By following [15], the penalty parameter $\gamma$ is searched in the range of 0.1 to 2.0 with an interval of 0.1, the regularization parameters $b_1$ and $b_2$ are tuned over $\{2^{-5},2^{-4},\cdots,2^{4},2^{5}\}$, and the number of fuzzy rules is searched in the range of 2 to 15 with an interval of 1. |
| 1-TSK-FC | All involved parameters are set to be the same as those in 0-TSK-FC. |
| S-KPC | The default settings are adopted for the parameter $\lambda$, kernel width $\sigma$, maximum number $H$ of rounds, threshold $\varepsilon$ and $\theta$. |
| 0-IS-TSK-FC | The number $R$ of fuzzy rules is tuned over $\{2,3,\cdots,14,15\}$, and the regularization parameter $\lambda$ is tuned over $\{10^{-5},10^{-4},\cdots,10^{4},10^{5}\}$. |
| 1-IS-TSK-FC | The number $R$ of fuzzy rules and regularization parameter $\lambda$ are set to be the same as those in 0-IS-TSK-FC. |

In our experiments, the number of intrinsic clusters in each dataset serves as the input for all the comparison algorithms except FRBC which can adaptively determine the number of clusters. To evaluate the clustering performance of all the comparison algorithms, two commonly used metrics, i.e., the clustering accuracy (ACC) and normalized mutual information (NMI) [31,56], are employed. Additionally, all comparison algorithms are implemented 10 times, and the average clustering performance is reported. All experiments are performed with MATLAB on a computer with an Intel(R) Xeon(R) CPU E5-2673 v4 with 2.30 GHz and 64 GB RAM.

### 4.2 Performance on Benchmark Datasets

An ideal interpretable style fuzzy clustering algorithm should not only solve clustering problems on datasets with clear styles but also serve as a feasible clustering tool regardless of whether the datasets have explicit styles. To examine the feasibility of IS-TSK-FC, fourteen benchmark datasets downloaded from the UCI [49] and KEEL [50] data repositories are used, and their details are summarized in Table 2.

**Table 2** Descriptions of the used fourteen benchmark datasets

| Names | #Samples | #Dimensions | #Clusters |
|---|---|---|---|
| Balance(*BAL*) | 625 | 4 | 3 |
| Lymphography(*LYM*) | 148 | 18 | 4 |
| Glass(*GLA*) | 214 | 9 | 6 |
| Haberman(*HAB*) | 306 | 3 | 2 |
| Heart(*HEA*) | 270 | 13 | 2 |
| Ionosphere(*ION*) | 351 | 33 | 2 |
| Newthyroid(*NEW*) | 215 | 5 | 3 |
| Flare(*FLA*) | 1066 | 11 | 6 |
| Wine(*WIN*) | 178 | 13 | 3 |
| Zoo(*ZOO*) | 101 | 16 | 7 |
| Wirelessindoorlocalization(*WIR*) | 2000 | 7 | 4 |
| Phoneme(*PHO*) | 5404 | 5 | 2 |
| Vehicle(*VEH*) | 846 | 18 | 4 |
| Yeast(*YEA*) | 1484 | 8 | 10 |

Tables 3 and 4 present the detailed clustering results of all the comparison algorithms on all the benchmark datasets. In Tables 3 and 4, "--" denotes values smaller than $10^{-4}$, and the best clustering results for each benchmark dataset are highlighted in bold. The following observations can be made.

1) Compared to the two basic clustering algorithms, FSC and FCM, the proposed IS-TSK-FC achieves better clustering results on most of the benchmark datasets. In addition, Tables 3 and 4 show that the ACC and NMI of IS-TSK-FC in most cases outperform those of the other state-of-the-art clustering algorithms. This may be because IS-TSK-FC considers the styles of data, which results in the fuzzy rules in IS-TSK-FC having powerful data representation capabilities.

**Table 3** Clustering performance (accuracy) of all comparison algorithms on all benchmark datasets (mean and standard deviation %)

| Datasets | FSC | FCM | FWCW_FCM | FRBC | 0-TSK-FC | 1-TSK-FC | S-KPC | 0-IS-TSK-FC | 1-IS-TSK-FC |
|---|---|---|---|---|---|---|---|---|---|
| *BAL* | 50.51±1.30 | 50.61±6.99 | 49.81±6.67 | 47.71±8.38 | 55.62±4.95 | 55.22±7.09 | **66.69**±17.00 | 55.02±1.22 | 57.41±8.43 |
| *LYM* | 58.92±5.00 | 41.22±0.00 | 56.01±5.98 | 53.04±4.07 | 42.36±6.80 | 46.01±3.29 | **62.03**±6.53 | 55.41±-- | 55.81±-- |
| *GLA* | 37.85±2.84 | 49.07±-- | 36.12±0.88 | 37.48±2.57 | 46.92±2.45 | 51.50±1.03 | 47.06±2.74 | 52.20±0.58 | **52.29**±0.15 |
| *HAB* | 74.51±0.00 | 50.98±0.00 | 73.20±0.00 | 74.25±0.99 | 71.67±8.68 | 74.93±1.53 | 74.28±0.94 | 76.97±0.17 | **77.03**±0.16 |
| *HEA* | **76.30**±-- | 59.26±-- | 58.19±5.46 | 55.63±4.72 | 66.96±4.81 | 62.81±4.33 | 63.33±-- | 66.59±0.23 | 66.96±0.16 |
| *ION* | 68.38±-- | 70.94±-- | 71.68±5.23 | 68.38±4.22 | 69.26±8.93 | 72.88±4.25 | 72.36±4.20 | **80.06**±-- | 77.52±3.59 |
| *NEW* | 85.12±-- | 79.33±0.64 | 62.74±3.82 | 76.93±6.64 | 84.05±5.81 | 83.40±1.08 | 80.23±1.06 | 91.67±7.41 | **96.28**±-- |
| *FLA* | 41.89±5.64 | 49.02±-- | 39.03±6.24 | 52.66±2.85 | 56.28±6.40 | 59.83±6.58 | 42.21±7.37 | 64.02±4.53 | **64.93**±7.87 |
| *WIN* | **93.26**±-- | 68.54±0.00 | 91.57±0.00 | 57.53±5.64 | 66.91±1.36 | 60.62±0.18 | 60.39±11.55 | 63.48±-- | 63.48±-- |
| *ZOO* | 72.57±7.98 | 58.88±7.38 | 62.77±7.77 | 58.61±5.07 | 74.55±7.88 | 75.84±1.88 | 67.03±6.48 | 77.62±1.16 | **84.16**±-- |
| *WIR* | 96.55±-- | 96.20±0.00 | 84.73±3.18 | 38.97±5.00 | 58.50±7.58 | 62.03±7.31 | 55.06±3.62 | 85.78±-- | **96.76**±-- |
| *PHO* | 61.21±0.00 | 68.34±0.00 | 63.10±0.00 | 47.67±8.37 | 71.44±1.81 | 70.63±-- | 74.45±0.52 | 70.65±0.00 | **77.89**±0.00 |
| *VEH* | **45.63**±0.00 | 37.12±-- | 45.39±-- | 38.06±3.07 | 35.79±3.20 | 36.23±5.11 | 37.55±1.18 | 41.62±2.81 | 42.39±3.21 |
| *YEA* | 33.42±1.29 | 32.42±0.42 | 33.71±2.23 | 37.41±2.56 | 32.90±3.73 | 34.47±3.39 | 34.87±0.22 | **39.80**±1.23 | 39.66±1.15 |

**Table 4** Clustering performance (normalized mutual information) of all comparison algorithms on all benchmark datasets (mean and standard deviation %)

| Datasets | FSC | FCM | FWCW_FCM | FRBC | 0-TSK-FC | 1-TSK-FC | S-KPC | 0-IS-TSK-FC | 1-IS-TSK-FC |
|---|---|---|---|---|---|---|---|---|---|
| *BAL* | 11.02±1.62 | 10.61±7.81 | 7.87±7.95 | 17.27±6.33 | 20.56±9.53 | 20.66±10.73 | **23.73**±18.84 | 17.25±1.68 | 20.76±19.68 |
| *LYM* | **20.89**±6.51 | 7.68±-- | 11.22±5.65 | 4.52±3.22 | 10.82±7.51 | 13.09±5.74 | 12.64±4.58 | 5.68±0.00 | 5.68±0.00 |
| *GLA* | 15.60±7.88 | 35.94±-- | 12.07±0.63 | 7.09±8.31 | 35.81±3.72 | 43.23±1.94 | 25.89±1.12 | 42.69±3.29 | **43.83**±2.46 |
| *HAB* | 6.36±-- | --±-- | 8.36±-- | 3.15±4.35 | 6.75±3.88 | 5.95±2.68 | 3.89±3.06 | 9.57±0.36 | **9.71**±0.33 |
| *HEA* | **20.51**±0.00 | 2.14±0.00 | 2.78±4.81 | 1.24±2.79 | 9.08±4.26 | 6.87±2.11 | 6.89±-- | 14.68±0.32 | 15.20±0.22 |
| *ION* | 11.97±-- | 12.99±-- | 18.21±12.57 | 13.18±11.39 | 11.67±11.51 | 16.12±8.49 | 11.66±4.90 | **35.78**±0.00 | 32.03±5.71 |
| *NEW* | 52.93±-- | 34.77±0.80 | 31.73±3.64 | 29.56±27.01 | 48.41±12.21 | 45.47±3.19 | 47.39±6.10 | 71.60±13.76 | **80.15**±-- |
| *FLA* | 32.54±5.69 | 51.38±-- | 21.53±8.95 | 34.00±4.49 | 44.74±4.82 | 48.34±5.47 | 24.16±8.67 | 54.83±0.39 | **54.91**±1.93 |
| *WIN* | **80.14**±-- | 41.68±-- | 73.91±-- | 44.66±2.64 | 43.60±2.39 | 26.34±15.06 |  | 28.37±0.00 | 28.37±0.00 |
| *ZOO* | 81.74±3.91 | 67.31±3.72 | 60.07±10.22 | 45.51±5.33 | 69.70±7.58 | 67.24±3.00 | 60.13±8.31 | 76.28±0.66 | **82.13**±-- |
| *WIR* | **90.63**±-- | 90.18±-- | 73.74±1.18 | 28.00±8.00 | 52.00±13.62 | 59.71±6.95 | 42.46±3.37 | 73.05±-- | 89.60±-- |
| *PHO* | 9.30±0.00 | 17.19±-- | 10.60±0.00 | 8.11±3.33 | 2.10±4.40 | 0.21±0.33 | 14.13±5.28 | 10.30±0.00 | **18.30**±0.00 |
| *VEH* | 19.20±-- | 9.86±-- | **19.78**±-- | 11.51±5.16 | 7.96±2.97 | 10.96±3.49 | 11.01±1.90 | 13.53±1.84 | 14.05±1.29 |
| *YEA* | 10.03±0.59 | 17.38±0.22 | 9.04±2.30 | 14.02±4.80 | 13.29±3.72 | 15.41±2.74 | 8.32±2.99 | 22.96±1.01 | **23.15**±1.07 |

2) IS-TSK-FC significantly outperforms FRBC on all the benchmark datasets. FRBC aims to transform the unsupervised learning task into a fuzzy rule-based supervised learning task [11]. This means that the clustering results of FRBC are obtained indirectly based on fuzzy inference on fuzzy rules, leading to the limited interpretability of FRBC. However, according to Algorithm 1, the clustering procedure of IS-TSK-FC is directly performed via fuzzy inference on fuzzy rules. Thus, the process of grouping the samples into their clusters using IS-TSK-FC is fully interpretable.

3) Considering the clustering algorithms directly based on fuzzy inference on fuzzy rules, IS-TSK-FC outperforms TSK-FC in most cases. In particular, compared with TSK-FC, the ACC and NMI of IS-TSK-FC are improved by approximately 7.18% and 19.66%, respectively, on *ION* and 34.73% and 29.89%, respectively, on *WIR*. This potentially indicates that the introduction of the styles of data effectively improves the clustering performance of IS-TSK-FC.

4) Compared to the S-KPC, which considers the styles of data [31], the IS-TSK-FC not only provides comprehensible clustering results based on fuzzy inference on fuzzy rules but also outperforms S-KPC on most of the benchmark datasets. This may be attributed to the fact that the fuzzy inference on the IF-parts of all fuzzy rules in IS-TSK-FC is equivalent to transforming the original feature space of the samples into a high-dimensional feature space [45], where the styles of different clusters are exploited to improve the discriminability of IS-TSK-FC for samples with different styles.

The results in Tables 3 and 4 indicate the superiority of the proposed IS-TSK-FC as an interpretable style fuzzy clustering method. Moreover, the results validate the feasibility of IS-TSK-FC as a basic clustering tool for solving clustering problems on the datasets with explicit styles of data or not.

To better understand IS-TSK-FC, we investigate how the number $R$ of fuzzy rules and the regularization parameter $\lambda$ influence the clustering performance of IS-TSK-FC. Here, due to the space limitations of this paper, we present experimental observations on six benchmark datasets that are randomly chosen from the fourteen benchmark datasets.

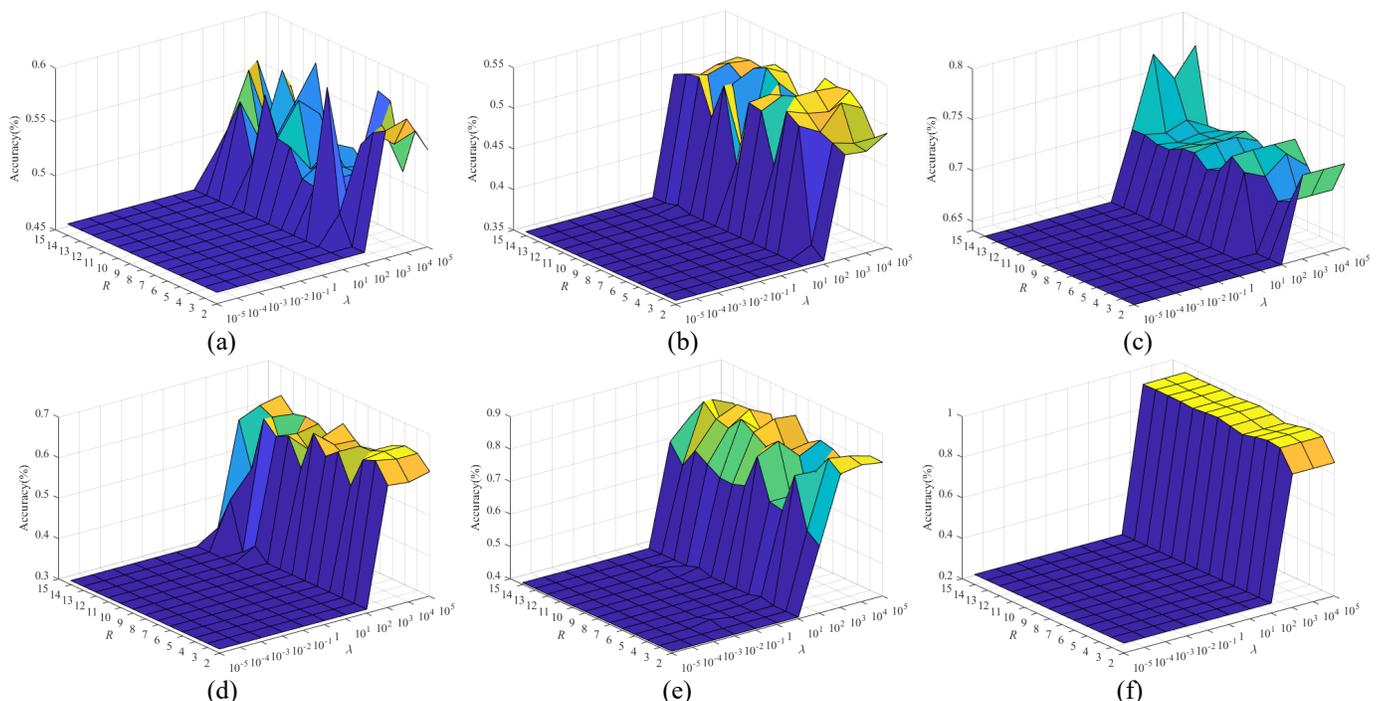

**Fig. 3** Visual illustration of the influence of different parameters in 1-IS-TSK-FC on the clustering accuracy on six benchmark datasets, where the parameters include the number $R$ of fuzzy rules and the regularization parameter $\lambda$. (a) *BAL*, (b) *GLA*, (c) *ION*, (d) *FLA*, (e) *ZOO*, and (f) *WIR*.

Fig. 3 shows the influence of the parameters $R$ and $\lambda$ on the clustering accuracy of 1-IS-TSK-FC on six benchmark datasets. According to Fig. 3, the following observations can be made.

1) The proposed IS-TSK-FC achieves the best clustering performance when the parameters $R$ and $\lambda$ are searched in $\{2, 3, \cdots, 14, 15\}$ and $\{10^{-5}, 10^{-4}, \cdots, 10^{4}, 10^{5}\}$, respectively, which are acceptable for tuning.

2) When $\lambda$ is small, the clustering performance of IS-TSK-FC remains essentially unchanged with different values of $R$. In this case, the overly flexible style information may be exploited for the six benchmark datasets, and the clustering performance of IS-TSK-FC is mainly dominated by the styles of data. Moreover, the clustering performance of IS-TSK-FC fluctuates when the value of $\lambda$ is relatively large for different values of $R$. This observation indicates that the style of the data in each cluster and the nuances between different styles can be appropriately captured through the regularization term in (21), which may account for the improved clustering performance of IS-TSK-FC.

3) The impact of varying $R$ and $\lambda$ with relatively large values of $\lambda$ is apparent based on the results for most of the benchmark datasets. This observation indicates that the clustering performance of IS-TSK-FC is sensitive to these two parameters. However, by considering observation 1) above, IS-TSK-FC can achieve optimal and robust clustering performance with narrow tuning ranges for these two parameters.

Since the alternating optimization strategy [19,38] is used to solve the consequent vector and style matrix suboptimization problems, which are the main components of the optimization problem of IS-TSK-FC in (21), the algorithm convergence is also investigated. Fig. 4 shows the convergence of the alternating optimization strategy when IS-TSK-FC achieves the optimal clustering performance on the six benchmark datasets used in Fig. 3. Fig. 4 demonstrates that the objective function of IS-TSK-FC in (21) reaches a stable value after several iterations. This observation indicates that when updating $\{\mathbf{P}_{gk}\}$ and $\{\mathbf{S}_k\}$ in an alternating manner, IS-TSK-FC can converge in a few steps.

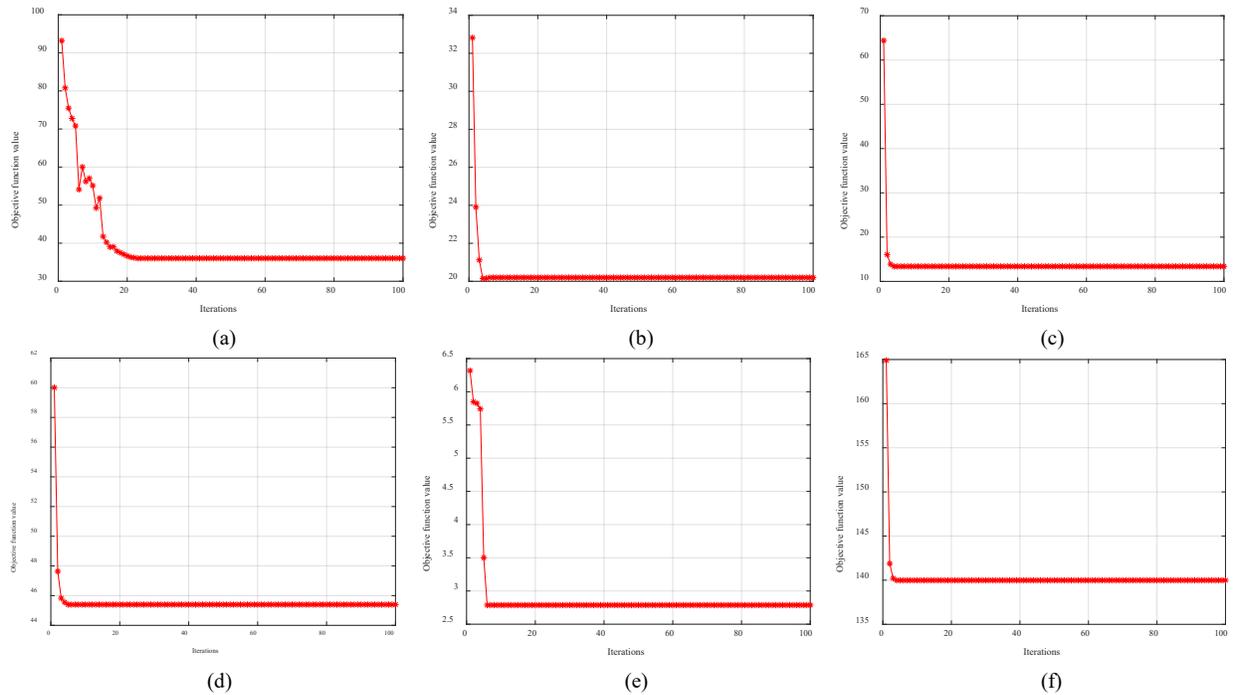

**Fig. 4** Convergence of the proposed IS-TSK-FC on the six benchmark datasets in Fig. 3. (a) *BAL*, (b) *GLA*, (c) *ION*, (d) *FLA*, (e) *ZOO*, and (f) *WIR*.

### 4.3 On Case Studies

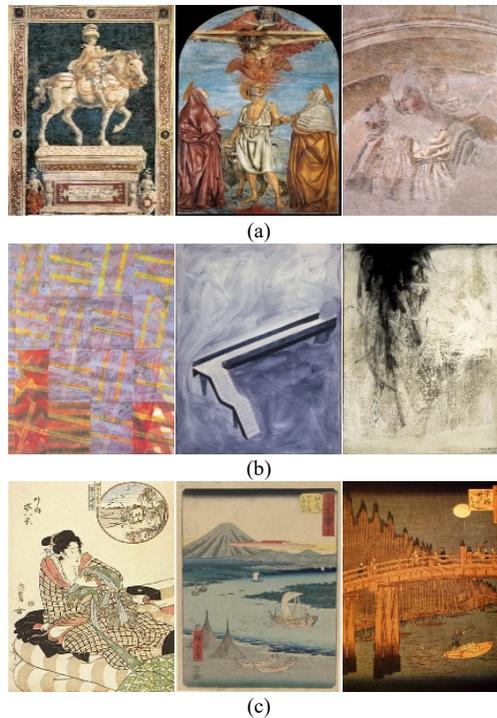

**Fig. 5** Three styles with comparable numbers of images in *WikiPainting*. (a) Early Renaissance, (b) Minimalism, and (c) Ukiyo-e.

In this subsection, the performance of all the comparison algorithms is evaluated in seven cases studies to assess the advantages of IS-TSK-FC for solving clustering problems on the datasets with explicit styles of data. The details of each dataset considered in the case studies are provided as follows.

1) The *WikiPainting* dataset is one of the most popular style recognition datasets and is commonly used for image style

classification [21-22]. This dataset consists of 85000 high-art images – mostly paintings – with 25 considerably different styles, such as Cubism and post Impressionism, as shown in Fig. 1(a). The dataset includes more than 1000 images for each style. However, *WikiPainting* is a highly imbalanced dataset, which can be determined by inspecting the number of images available for each of the 25 styles [21-22]. For example, 13060 images are available for the Impressionism style, while only 1279 images are available for the Mannerism style. Here, according to [21-22], three styles with comparable numbers of images, i.e., Early Renaissance, Minimalism, and Ukiyo-e, as shown in Fig. 5, are randomly chosen to form a simplified *WikiPainting* (abbr. *SimWikiPainting*) dataset with 3895 images. In addition, following [21-22], the original images are first resized to 256 × 256 pixels, and the features are subsequently reduced to the first 50 principal components via principal component analysis (PCA) by following [19].

2) The *Hipster Wars* dataset is a benchmark dataset for fashion style recognition [23-24] collected from an online competitive style rating game. In this dataset, there are 1893 clothing outfit images that can be divided into 5 style categories, hipster, bohemian, pinup, preppy, and goth, as shown in Fig. 6. Following [19,24], kernel principal component analysis (KPCA) is employed to project all images in *Hipster Wars* onto the first 25 principal components.

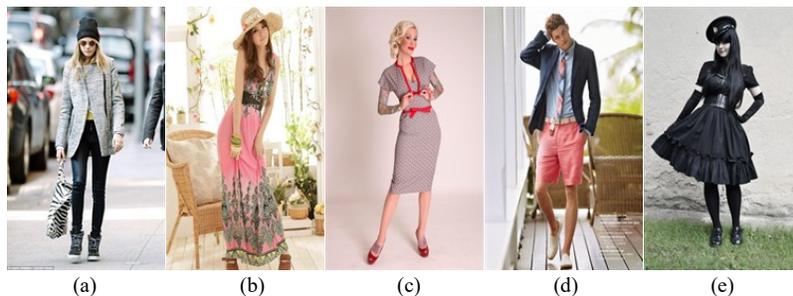

(a)　　　　　(b)　　　　　(c)　　　　　(d)　　　　　(e)

**Fig. 6** Examples of clothing outfit images associated with 5 styles in *Hipster Wars* dataset. (a) Hipster, (b) bohemian, (c) pinup, (d) preppy, and (e) goth.

3) The architecture style dataset is the largest publicly available dataset collected from Wikimedia and is commonly used for architecture style recognition [24-25]. This dataset includes 4794 architectural images that can be divided into 25 diverse style categories. Fig. 7 shows examples of partial architectural images for 8 style categories. In our experiments, the dataset with 10 style categories in [25] is adopted for the clustering task, and we term this dataset *ASD10* for simplicity. For a fair comparison, the Histogram of Oriented Gradient (HOG) descriptors are first considered to describe the architectural images in both *ASD10* and *ASD25*, and then PCA is employed for each HOG feature following [24-25].

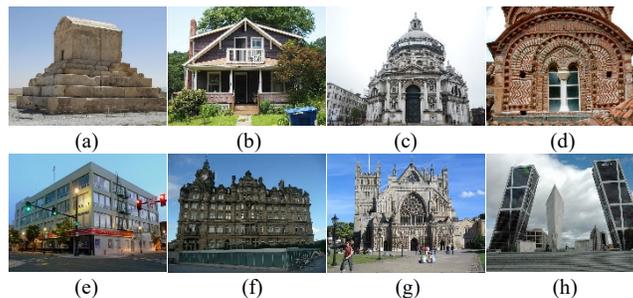

(a)　　　　　(b)　　　　　(c)　　　　　(d)

(e)　　　　　(f)　　　　　(g)　　　　　(h)

**Fig. 7** Examples of partial architectural images with 8 style categories. (a) Achaemenid, (b) American craftsman, (c) Baroque, (d) Byzantine, (e) Chicago School, (f) Edwardian, (g) Gothic, and (h) Postmodern.

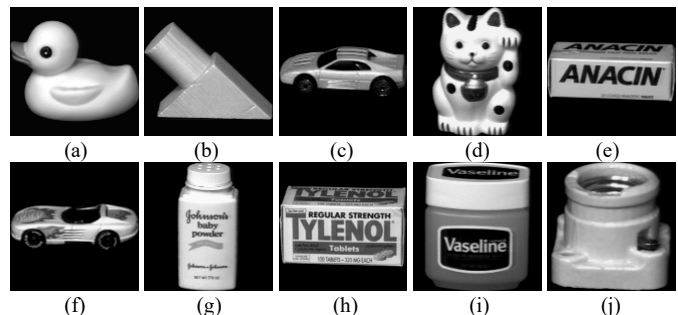

(a)　　　(b)　　　(c)　　　(d)　　　(e)

(f)　　　(g)　　　(h)　　　(i)　　　(j)

**Fig. 8** Examples of the first images of eight objects in *COIL-20*.

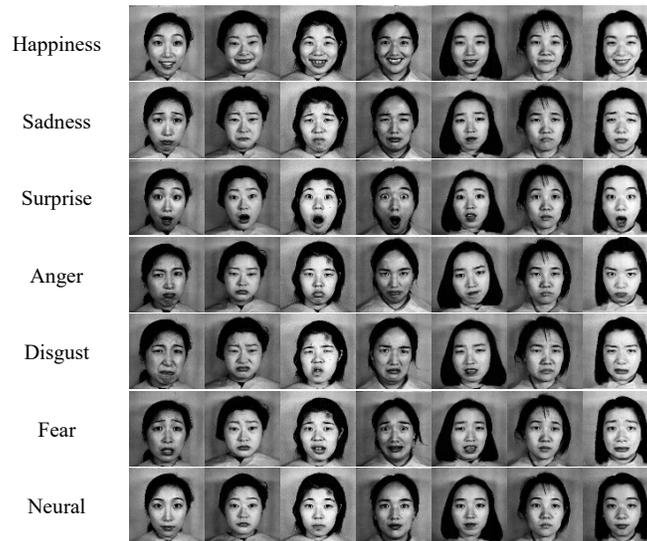

**Fig. 9** Examples of partial images of each basic facial expression in *JAFFE*.

4) The processed Columbia object image library *COIL-20* dataset contains 1440 grayscale images of 20 objects. To collect the data, each object was placed on a motorized turntable with a black background. Then, the turntable was rotated 360 degrees to vary the object pose with a fixed camera, and 72 images of each object were collected [51]. Fig. 8 shows the first image for each object in *COIL-20*. According to [52], the composition of a photograph is closely associated with its object. Thus, the same object is typically photographed in a similar style. In our experiments, we considered images of 8 objects, i.e., the first, fourth, fifth, eighth, thirteenth, fifteenth, eighteenth, and nineteenth objects, are randomly chosen to generate a simplified *COIL-20* dataset (abbr. *SimCOIL-20*). By following [19], KPCA is applied to project all images onto the first 25 principal components before clustering.

**Table 5** Details of the *EEG* dataset

| Subjects | Subsets | #Sizes | Descriptions of each subset |
|---|---|---|---|
| Healthy | A | 100 | EEG signals collected from healthy crowds with eyes open. |
| | B | 100 | EEG signals collected from healthy crowds with eyes closed. |
| Epileptic | C | 100 | Epileptic EEG signals collected from the hippocampal formation of the opposite hemisphere of the brain during seizure free intervals. |
| | D | 100 | Epileptic EEG signals collected from the epileptogenic zone during seizure free intervals. |
| | E | 100 | Epileptic EEG signals collected during happening seizure activity. |

5) The Japanese female facial expression (*JAFFE*) database is widely used for facial expression recognition [35,53]. It consists of 213 facial expression images obtained by 10 expressers performing different poses. Each image is 256 × 256 pixels in size and corresponds to one of seven basic facial expressions: happiness, sadness, surprise, anger, disgust, fear, or neural. Besides, three or four images for each basic facial expression are available for each expresser. Fig. 9 shows examples of partial images for each basic facial expression. Fig. 9 shows that the images of each expression present the same emotion which indicates that the images of the seven basic facial expressions have different expression styles [35]. According to [19], KPCA is used to project all images in *JAFFE* onto the first 15 principal components before clustering.

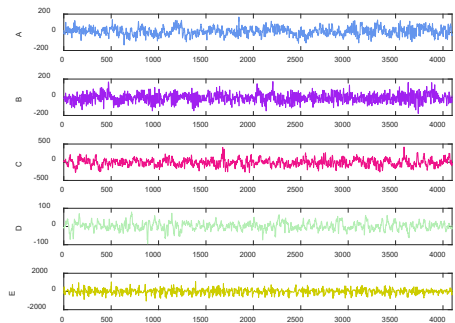

**Fig. 10** Visualization of EEG segments in the five subjects in the *EEG* dataset.

6) The cardiac single-photon emission computed tomography (SPECT) database is commonly used in hear abnormality detection [54]. This dataset includes 267 SPECT images that can be classified into normal and abnormal categories. According to [54], SPECT images present two unique styles of information for patients with normal and abnormal heart diagnosis. In our experiments, the *SPECTF* dataset derived from the SPECT database is used. In the *SPECTF* dataset, each sample consists of 44 continuous feature patterns describing the corresponding patient [54].

7) The epileptic electroencephalogram (*EEG*) dataset [55] is widely used epilepsy detection, which is a common neurological disorder. In this dataset, there are five subsets (termed A, B, C, D, and E) of EEG signals. In addition, each subset contains 100 single-channel EEG segments, including 4096 samples during an EEG time series (23.6s duration). The details of the *EEG* dataset are summarized in Table 5, and five segments of the original EEG data in the five subsets are visualized in Fig. 10.

Fig. 10 shows that the EEG signals of each subset have distinct waveforms. For example, the waveforms of EEG signals of healthy people are clearly different from those of EEG signals of epileptic people. Moreover, for the same group, differences can be observed among the waveforms of the EEG signals of different subjects, e.g., subjects C and E. These observations indicate that the EEG signals of the subjects in the *EEG* dataset have explicit styles. Since detecting epilepsy based on the original EEG signals in the *EEG* dataset cannot achieve desirable performance [55], KPCA is applied to project the original EEG signals onto the first 10 principal components by following [19].

**Table 6** Detailed clustering performance of all the comparison algorithms on ten datasets for the eight case studies in terms of ACC (%)

| Datasets | FSC | FCM | FWCW_FCM | FRBC | 0-TSK-FC | 1-TSK-FC | S-KPC | 0-IS-TSK-FC | 1-IS-TSK-FC |
|---|---|---|---|---|---|---|---|---|---|
| *SimWikiPainting* | 36.07±0.71 | 53.27±0.00 | 49.17±5.00 | 38.31±3.99 | 48.61±2.89 | 49.30±4.05 | 47.03±6.02 | 53.63±-- | **53.71**±-- |
| *Hipster Wars* | 28.01±0.79 | 37.28±4.25 | 37.47±0.54 | 27.24±2.73 | 27.86±3.53 | 36.24±0.00 | **43.22**±4.24 | 39.11±0.58 | 37.11±0.74 |
| *ASD10* | 13.90±0.70 | 20.57±0.95 | 21.12±1.49 | 20.89±1.56 | 19.17±1.51 | 22.50±1.69 | **25.58**±0.65 | 19.90±1.58 | 22.56±-- |
| *SimCOIL-20* | 22.27±2.11 | 23.81±0.73 | 21.96±-- | 16.55±3.00 | 19.95±3.50 | 25.00±0.00 | **29.79**±2.43 | 27.62±0.72 | 27.71±0.82 |
| *JAFFE* | 27.37±2.25 | 27.92±1.38 | 24.04±3.24 | 25.75±3.96 | 26.43±1.52 | 36.15±0.00 | 32.77±1.73 | 36.76±-- | **39.01**±1.23 |
| *SPECT* | 53.88±-- | 57.67±1.53 | 55.96±3.20 | 56.24±7.36 | 61.63±2.99 | 62.82±2.47 | 60.49±4.13 | 62.04±0.77 | **63.67**±0.00 |
| *SPECTF* | 66.29±3.98 | 62.55±0.00 | 61.54±1.82 | 65.54±6.59 | 68.05±8.45 | 68.20±6.36 | 64.42±4.77 | **68.58**±4.55 | 68.31±2.16 |
| *EEG* | 33.74±3.39 | 33.46±0.44 | 34.06±0.84 | 29.58±3.15 | 31.76±3.82 | 41.04±4.61 | 35.62±3.50 | 38.80±-- | **43.80**±-- |

**Table 7** Detailed clustering performance of all the comparison algorithms on ten datasets for the eight case studies in terms of NMI (%)

| Datasets | FSC | FCM | FWCW_FCM | FRBC | 0-TSK-FC | 1-TSK-FC | S-KPC | 0-IS-TSK-FC | 1-IS-TSK-FC |
|---|---|---|---|---|---|---|---|---|---|
| *SimWikiPainting* | 0.25±0.11 | 11.29±-- | 8.99±3.22 | 1.76±1.59 | 7.15±2.04 | 9.08±0.98 | 6.89±3.68 | 12.31±0.01 | **12.36**±0.00 |
| *Hipster Wars* | 3.58±0.62 | 16.11±3.29 | 10.01±0.38 | 2.89±2.09 | 4.14±4.16 | 19.99±10.02 | **21.06**±6.13 | 20.77±0.39 | 20.98±1.87 |
| *ASD10* | 1.66±0.33 | 7.03±1.80 | 7.59±0.73 | 2.61±1.46 | 3.68±2.16 | 6.03±5.64 | 7.41±0.42 | 7.23±2.60 | **11.34**±-- |
| *SimCOIL-20* | 8.09±2.36 | 18.23±0.72 | 11.76±-- | 3.71±2.44 | 7.69±3.3 | 9.85±4.82 | **29.46**±1.58 | 26.63±4.18 | 26.63±0.30 |
| *JAFFE* | 10.35±2.60 | 15.02±1.23 | 7.96±2.89 | 4.66±2.19 | 10.05±2.11 | 25.91±0.00 | 19.21±1.32 | 20.18±-- | **26.38**±1.11 |
| *SPECT* | 3.41±0.18 | 1.54±0.76 | 3.45±2.23 | 4.18±2.22 | 5.39±3.04 | 7.15±2.55 | 5.05±2.06 | **9.78**±2.35 | 4.63±-- |
| *SPECTF* | 5.47±2.54 | **8.98**±0.00 | 5.18±2.51 | 2.12±2.32 | 3.51±2.78 | 5.27±2.96 | 6.16±2.69 | 6.34±2.00 | 6.48±0.94 |
| *EEG* | 10.29±1.91 | 12.78±0.25 | 12.15±0.63 | 6.56±2.41 | 7.78±4.57 | 18.53±3.62 | 10.33±3.27 | 20.91±0.12 | **24.01**±0.00 |

**Table 8** Ranks of each comparison algorithm on the datasets in all case studies in terms of ACC in Table 6

| Datasets | FSC | FCM | FWCW_FCM | FRBC | 0-TSK-FC | 1-TSK-FC | S-KPC | 0-IS-TSK-FC | 1-IS-TSK-FC |
|---|---|---|---|---|---|---|---|---|---|
| *SimWikiPainting* | 9 | 3 | 5 | 8 | 6 | 4 | 7 | 2 | 1 |
| *Hipster Wars* | 7 | 4 | 3 | 9 | 8 | 6 | 1 | 2 | 5 |
| *ASD10* | 9 | 6 | 4 | 5 | 8 | 3 | 1 | 7 | 2 |
| *SimCOIL-20* | 6 | 5 | 7 | 9 | 8 | 4 | 1 | 3 | 2 |
| *JAFFE* | 6 | 5 | 9 | 8 | 7 | 3 | 4 | 2 | 1 |
| *SPECT* | 9 | 6 | 8 | 7 | 4 | 2 | 5 | 3 | 1 |
| *SPECTF* | 5 | 8 | 9 | 6 | 4 | 3 | 7 | 1 | 2 |
| *EEG* | 6 | 7 | 5 | 9 | 8 | 2 | 4 | 3 | 1 |
| *AveRank* | 7.13 | 5.50 | 6.25 | 7.63 | 6.63 | 3.38 | 3.75 | 2.88 | 1.88 |

\**AveRank* denotes the average rank of a comparison algorithm on all datasets in the case studies.

Tables 6 and 7 present the clustering results of all the comparison algorithms on the datasets with explicit styles of data in all case studies. We observe that S-KPC and IS-TSK-FC, which consider the styles of data to perform clustering, outperform most of the other comparison algorithms on all datasets in the case studies. In addition, compared to TSK-FC, which also provides comprehensible clustering results via direct fuzzy inference on fuzzy rules [15], IS-TSK-FC obtains better clustering performance on all datasets in the case studies. These observations demonstrate that considering the styles of data may be an effective approach to discriminate samples with different styles during clustering. Moreover, S-KPC has superior clustering performance on *Hipster Wars*, *ASD10* and *SimCOIL-20*, which may attribute to the consideration of the twofold data representations, i.e., style matrices for representing style information of clusters and enhanced nodes for improved data discrimination [31]. However, IS-TSK-FC can produce fully interpretable clustering results because its clustering procedure is performed via direct fuzzy inference on fuzzy rules.

The clustering results in Tables 6 and 7 demonstrate the feasibility of IS-TSK-FC on the datasets with obvious styles of data. According to these clustering results, we next investigate the performance of IS-TSK-FC through statistical test. To this end, a robust and nonparametric statistical test method [61] is employed, which is briefly described as follows, and its implementation details can be found in [15,31,35,61].

First, in the statistical test method, we assume that all the comparison algorithms have identical performance on all datasets. However, whether this assumption holds based on the Friedman test [61] needs to be checked. When conducting the Friedman test, we need to calculate the statistical value $F_F$, and the initial assumption should be rejected if $F_F > F\big((N_c-1),(N_c-1)(N_d-1)\big)$, where $N_c$ and $N_d$ denote the numbers of comparison algorithms and datasets, respectively. If the initial assumption is rejected, we implement the Bonferroni-Dunn test [61], where the critical value $CD$ is determined as $CD = q_\alpha\sqrt{N_c(N_c+1)/6N_d}$. According to [61], the significance level $\alpha$ is usually set to 0.05, and the specific value of $q_\alpha$ can be found in the table of critical values for post-hoc tests after the Friedman test in [61]. We rank all the comparison algorithms based on their clustering accuracies on all datasets in the case studies shown in Table 8 and conclude that there is a significant difference in the performance of two comparison algorithms if the difference between their average ranks is larger than $CD$ [61].

According to Table 6, the numbers of comparison algorithms and datasets are $N_c = 9$ and $N_d = 8$, respectively. Moreover, based on the average ranks of all comparison algorithms on all datasets in the case studies, as shown in Table 8, the statistical value is $F_F \approx 9.56$ and $F(8,56) \approx 2.11$ [61]. Consequently, the above assumption that all comparison algorithms perform the same on all datasets in the case studies is rejected because $F_F > F(8,56)$. In addition, since $q_{0.05} = 2.724$, the critical value is $CD \approx 3.73$. Compared to the other comparison algorithms, 1-IS-TSK-FC and 0-IS-TSK-FC rank first and second on all datasets, which is consistent with the results shown in Tables 6 and 7. Moreover, 1-IS-TSK-FC shows significantly different performance with respect to FSC, FWCW_FCM, FRBC and 0-TSK-FC because the difference values between the average ranks of 1-IS-TSK-FC and these four comparison algorithms are 5.25, 4.37, 5.75 and 4.75, respectively, which are larger than the critical value $CD$. The same result can be obtained according to the difference values between the average ranks of 0-IS-TSK-FC and FSC, FRBC, and 0-TSK-FC. These results demonstrate the superiority of IS-TSK-FC in performing clustering on the datasets with explicit styles of data. Please note that the same statistical results can be achieved if all comparison algorithms are ranked based on the clustering performance in terms of NMI, as shown in Table 7.

**Table 9** Average running time (seconds) of TSK-FC and IS-TSK-FC on the datasets in all case studies

| Datasets | 0-TSK-FC | 1-TSK-FC | 0-IS-TSK-FC | 1-IS-TSK-FC |
|---|---|---|---|---|
| *SimWikiPainting* | 86.33 | 176.02 | 1.17 | 22.30 |
| *Hipster Wars* | 25.41 | 35.89 | 0.18 | 10.25 |
| *ASD10* | 71.49 | 126.65 | 0.30 | 41.76 |
| *SimCOIL-20* | 3.09 | 6.54 | 0.11 | 0.89 |
| *JAFFE* | 0.70 | 1.80 | 0.03 | 0.26 |
| *SPECT* | 0.16 | 0.45 | 0.03 | 0.24 |
| *SPECTF* | 0.09 | 0.66 | 0.04 | 0.75 |
| *EEG* | 1.57 | 2.91 | 0.07 | 0.17 |

Since the clustering procedures of both TSK-FC and IS-TSK-FC are fully dominated by direct fuzzy inference on fuzzy rules, we also tabulate the average running time of these two algorithms on the datasets in all case studies, as shown in Table 9. For a fair comparison, the same number of fuzzy rules is used in both TSK-FC and IS-TSK-FC, and this value is set to 5 in our experiments. According to Table 9, IS-TSK-FC clearly has a shorter running time than TSK-FC on all datasets in the case studies, especially *SimWikiPainting*, *Hipster Wars* and *ASD10*, where the numbers of samples in these three datasets (i.e., 3895, 1893 and 2017, respectively) are relatively large. This observation is consistent with the computational complexity analysis of IS-TSK-FC in Subsection III.C. Based on Tables 6, 7, 8 and 9, we can conclude that the proposed IS-TSK-FC, which considers the styles of data in interpretable cluster analysis, can solve clustering problems more efficiently than TSK-FC on the datasets with explicit styles of data.

## 5 Conclusion

In this paper, an interpretable style TSK fuzzy clustering algorithm, IS-TSK-FC, which aims to generate comprehensible clusters by considering the styles of data, is proposed. Due to the introduction of the styles of data, a novel type of TSK fuzzy rule, in which the consequent parts include not only conventional consequent parameters (as seen in (1)) but also the style matrices (as seen in (15)), is formed. This allows IS-TSK-FC to have a powerful data representation capability. Besides, once the antecedents of all fuzzy rules are determined, the consequents of all fuzzy rules can be optimized in an alternating manner. In particular, each cluster generated by IS-TSK-FC corresponds to its own consequent vector rather than all generated clusters corresponding to only one consequent vector of all fuzzy rules. Thus, the clustering procedure of IS-TSK-FC via direct fuzzy inference on the new TSK fuzzy rules is transparent. The clustering results on the benchmark datasets validate the effectiveness of IS-TSK-FC as a basic tool for solving clustering problems on the datasets with or without explicit styles of data. Moreover, the clustering results in the case studies, especially the statistical test results, demonstrate the superiority of IS-TSK-FC for solving clustering problems on the datasets with explicit styles of data.

In the future, we will continue to explore two main aspects of IS-TSK-FC. First, although the experiments on the datasets in the case studies demonstrate the superiority of IS-TSK-FC, the nuances between different styles of data cannot be explicitly defined by the regularization term in problem (21). We will attempt to design a more interpretable regularization term in future work. Second, as shown in (36), when the dimensionality of the samples is high or the number of fuzzy rules is large, the dimension of the style matrix is high, which can increase the computational complexity of the optimization problem of IS-TSK-FC. Thus, the style matrix in IS-TSK-FC will be investigated with deep learning models, such as the consensus style centralizing auto-encoder in [24], to more efficiently learn style feature representations of data.

**Funding** This work was supported in part by the National Natural Science Foundation of China under Grants 62106025 and 51877013, in part by the Natural Science Foundation of Jiangsu Province under Grant BK20210940, in part by Suzhou Municipal Science and Technology Plan Project under Grant SYG202351, in part by Zhangjiagang Municipal Science and Technology Plan Project under Grant ZKYY2222, and in part by Qinglan Project of Jiangsu Province.

**Code availability** https://github.com/gusuhang10/IS-TSK-FC

**Declarations**

**Conflict of interest** The authors declare that they have no known competing financial interests or personal relationships that could have appeared to influence the work reported in this paper.
**Ethical approval** The article has the approval of all the authors.